\documentclass[preprint,12pt]{elsarticle}

\usepackage{amssymb}
\usepackage{amsmath}
\usepackage{graphicx}

\usepackage{geometry}
\usepackage[colorlinks=true]{hyperref}
\usepackage{orcidlink} 
\usepackage{lineno}
\usepackage{subcaption}
\usepackage{hyperref}
\usepackage{colortbl}
\definecolor{rowgray}{gray}{0.95}
\usepackage{tabulary}
\usepackage{booktabs,array,multirow}
\usepackage{hyperref}

\usepackage{makecell}

\journal{International Journal of Electronics and Communications}

\begin{document}
\begin{frontmatter}

\thispagestyle{empty}
\begin{center}
	\vspace*{4cm}
	{\Large \textbf{Accepted Manuscript Notice}} \\[1.5cm]
	
	This is the \textbf{accepted version} of the manuscript published in: \\[0.5cm]
	
	\textit{AEU - International Journal of Electronics and Communications} \\[0.5cm]
	
	DOI: \href{https://doi.org/10.1016/j.aeue.2025.156003}{https://doi.org/10.1016/j.aeue.2025.156003} \\[2cm]
	
	© 2025 Elsevier Ltd. All rights reserved.
\end{center}
\newpage



\title{AI-Powered Inverse Design of Ku-Band SIW Resonant Structures by Iterative Residual Correction Network}

\author[inst1]{\texorpdfstring{Mohammad Mashayekhi\,\orcidlink{0009-0004-4865-1264}\corref{cor1}}{Mohammad Mashayekhi}}
\cortext[cor1]{Corresponding author.}
\ead{masham7691@gmail.com}

\author[inst2]{\texorpdfstring{Kamran Salehian\,\orcidlink{0009-0008-5411-7278}}{Kamran Salehian}}

\author[inst1]{\texorpdfstring{Abbas Ozgoli}{Abbas Ozgoli}}

\author[inst3]{\texorpdfstring{Saeed Abdollahi\,\orcidlink{0009-0000-8157-7001}}{Saeed Abdollahi}}

\author[inst2]{\texorpdfstring{Abdolali Abdipour\,\orcidlink{0000-0002-0194-1983}}{Abdolali Abdipour}}

\author[inst4]{\texorpdfstring{Ahmed A. Kishk\,\orcidlink{0000-0001-9265-7269}}{Ahmed A. Kishk}}

\affiliation[inst1]{
  organization={Department of Electrical Engineering, Iran University of Science and Technology (IUST)},
  city={Tehran},
  country={Iran}
}

\affiliation[inst2]{
  organization={Department of Electrical Engineering, Amirkabir University of Technology (Tehran Polytechnic)},
  city={Tehran},
  country={Iran}
}

\affiliation[inst3]{
  organization={Department of Electrical and Computer Engineering, Isfahan University of Technology (IUT)},
  city={Isfahan},
  country={Iran}
}

\affiliation[inst4]{
  organization={Department of Electrical and Computer Engineering, Concordia University},
  city={Montreal},
  country={Canada}
}

\begin{abstract}
Designing high-performance substrate-integrated waveguide (SIW) filters with both closely spaced and widely separated resonances is challenging. Consequently, there is a growing need for robust methods that reduce reliance on time-consuming electromagnetic (EM) simulations.

In this study, a deep learning-based framework was developed and validated for the inverse design of multi-mode SIW filters with both closely spaced and widely separated resonances. A series of SIW filters were designed, fabricated, and experimentally evaluated. A three-stage deep learning framework was implemented, consisting of a Feedforward Inverse Model (FIM), a Hybrid Inverse-Forward Residual Refinement Network (HiFR\textsuperscript{2}-Net), and an Iterative Residual Correction Network (IRC-Net).

The design methodology and performance of each model were systematically analyzed. Notably, IRC-Net outperformed both FIM and HiFR\textsuperscript{2}-Net, achieving systematic error reduction over five correction iterations. Experimental results showed a reduction in mean squared error (MSE) from 0.00191 to 0.00146 and mean absolute error (MAE) from 0.0262 to 0.0209, indicating improved accuracy and convergence.

The proposed framework demonstrates the capability to enable robust, accurate, and generalizable inverse design of complex microwave filters with minimal simulation cost. This approach is expected to facilitate rapid prototyping of advanced filter designs and could extend to other high-frequency components in microwave and millimeter-wave technologies.

\end{abstract}

\begin{graphicalabstract}
\centering
\includegraphics[width=0.7\textwidth]{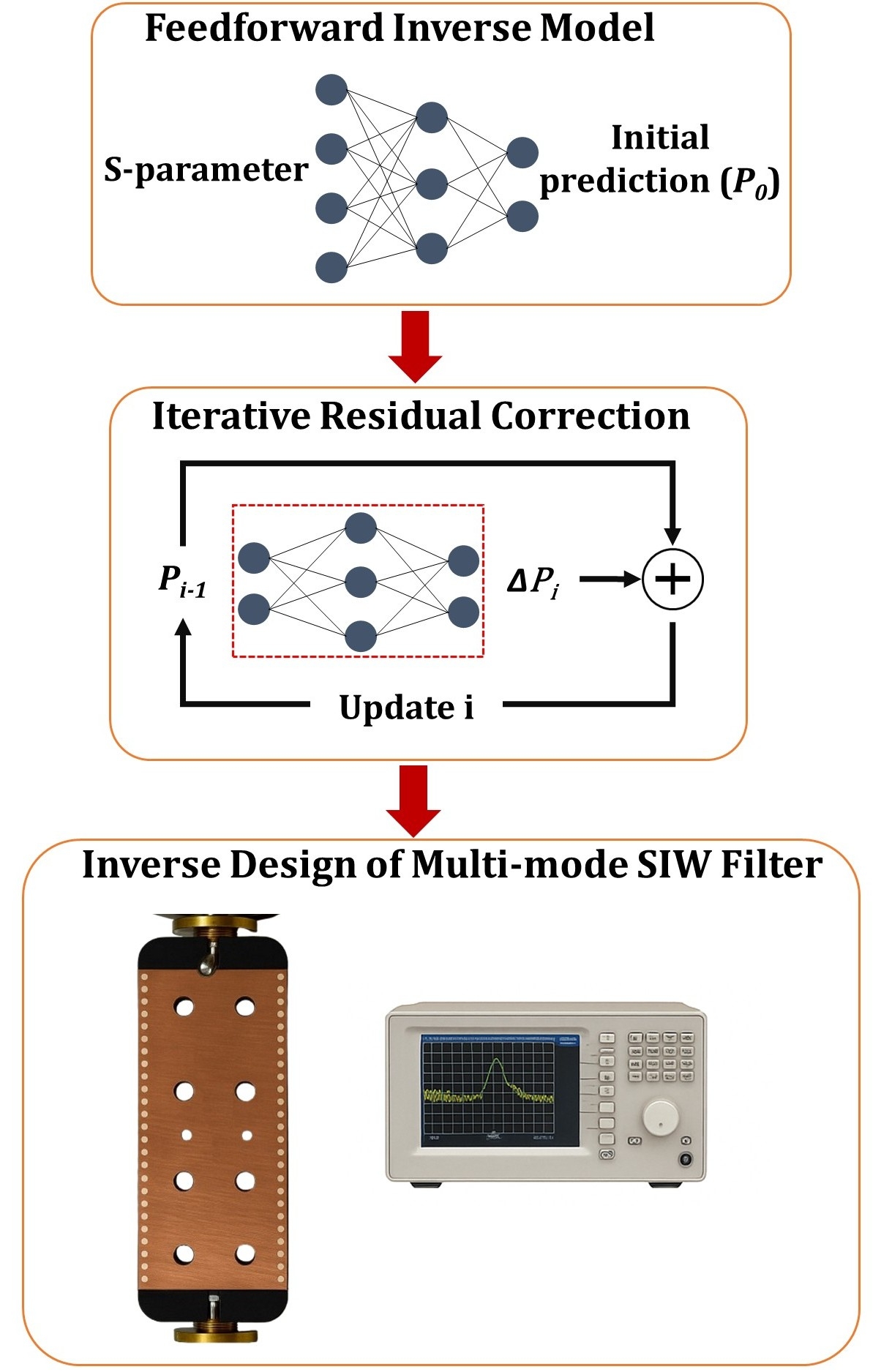}
\end{graphicalabstract}

\begin{highlights}
\item Novel SIW filters with both close and wideband resonances are realized
\item Deep learning trained on large SIW dataset enables accurate inverse design
\item HiFR$^2$-Net unifies inverse, forward, and residual learning for refinement
\item IRC-Net applies iterative correction to reduce design estimation error
\item Experimental results confirm high precision and strong real-world performance
\end{highlights}


\end{frontmatter}


\section{Introduction}
Resonant microwave structures underpin many modern filter designs, offering high quality-factor responses through carefully engineered cavities and resonators. Microwave filters play a crucial role in suppressing unwanted frequencies while preserving the integrity of the signal. The precise design of these filters is essential to ensure effective channelization and maintain communication quality \cite{10486926,10417067,ZAKHAROV2024155131}. Among the various classes of filters, multimode filters are particularly significant due to their compact size and ability to support multiple frequency bands, meeting the requirements and demands of modern communication systems, especially multi-standard wireless communication\cite{7054566,s22051961}.

Multimode microwave filters have been designed and implemented across a range of frequency bands \cite{electronics10222853}. However, the Ku-band stands out due to its growing importance in supporting high-capacity, data-intensive applications such as HD video streaming, real-time data transmission, and broadband satellite Internet. The growing use of LEO and geostationary satellites increases the need for high-performance components in this frequency range. To address demands like compact size, integration, and high selectivity, various designs have been proposed. For example, \cite{7859469} presented a planar bandpass filter using stepped-impedance resonators and multiple coupling paths. It offers a good balance between size, performance, and cost for PCB-based systems. In another study, \cite{10107477} analyzed the Starlink Ku-band downlink, highlighting the importance of filters in ensuring stable broadband in dense satellite networks. Consequently, the development of compact and efficient multimode filters operating in the Ku-band has become a critical objective in the design of next-generation communication systems.

While multimode resonant structures have been implemented using technologies such as microstrip~\cite{10896684} and conventional waveguides~\cite{10038507}, SIW technology offers a compelling alternative for high-frequency applications, particularly in the Ku-band. By embedding rows of conductive vias into a dielectric substrate, SIW preserves the low loss and high quality factor of traditional waveguides while enabling planar fabrication, compact size, and cost-effective integration~\cite{FARASHAHI201815}. These features make SIW well-suited for modern microwave systems, allowing seamless integration with antennas, amplifiers, and other RF components. In this context, \cite{10648671} investigated interconnect solutions for SIW structures at millimeter-wave frequencies, addressing key challenges in compact integration. Additionally, \cite{Darling2025} reviewed recent progress in SIW-based antenna technologies, highlighting the growing role of reconfigurable designs and AI-assisted methods in advancing 5G and MIMO systems. Together, these efforts demonstrate SIW’s potential as an effective platform for compact, efficient, and high-performance multimode resonators used in Ku-band single- and multi-band microwave filters.

Traditional design methods for electromagnetic structures—including antenna arrays and microwave components—typically depend on iterative full-wave simulations, heuristic optimizations, and labor-intensive, expert-driven trial-and-error approaches. While accurate, these methods are computationally intensive and time-consuming, particularly for high-dimensional and nonlinear design spaces. In recent years, data-driven approaches, particularly deep learning (DL), have emerged as powerful alternatives. The DL has shown promise in both forward modeling and inverse design. Numerous studies have demonstrated the effectiveness of DL-based methods in a wide range of electromagnetic applications, including phase retrieval in phased antenna arrays \cite{9760077}, pulse width and delay prediction in time-modulated antenna arrays for multibeam synthesis \cite{Mashayekhi2024}, automatic metasurface synthesis over broad frequency bands \cite{Ghorbani2021,Ghorbani2021b}, inverse design of reconfigurable and multiband antennas using deep learning frameworks \cite{Mashayekhi2023,10130105}, electromagnetic inverse scattering \cite{10702499}, microwave biomedical diagnostics using S-parameter profiles instead of imaging data \cite{Beyraghi2023}, and fast and accurate surrogate modeling of antennas using convolutional neural networks to predict near- and far-field parameters with significant speedup over full-wave simulations \cite{10472507}.

Deep learning approaches to microwave component design generally fall into two categories: Forward modeling and inverse design. Forward modeling networks take device geometry as input and predict its scattering response (e.g., S-parameters). In \cite{USTUN201954}, a deep neural network (DNN) is employed to predict the resonant frequency of E-shaped patch antennas (ESPAs) with high accuracy. The model is trained on a dataset of 144 simulated ESPAs and validated experimentally, demonstrating the potential of DNNs for rapid antenna analysis. Similarly, \cite{10944803} introduces an attention-based deep neural network combining convolutional neural network, joint channel–spatial attention, and residual connections that have been proposed for forward modeling of spatio-temporal modulated (STM) non-reciprocal microwave circuits, aiming to improve prediction accuracy and generalization across various circuit configurations.  

Inverse design methods, in contrast, aim to infer a geometry for a given desired response. Early examples include Kabir et al., who introduced neural-network inverse modeling for microwave filters, formulating the problem where the inputs are desired electrical specifications and the outputs are filter dimensions \cite{4470587}. Since then, researchers have tried to overcome implementation challenges and reduce errors in model parameter extraction. In \cite{8798884}, a single-stage neural network utilizing smooth ReLU and sigmoid activations is proposed for inverse modeling and parameter extraction of microwave filters. However, the model exhibits signs of overfitting, is trained on only 41 frequency samples for S-parameters, and lacks experimental validation, limiting its robustness and practical applicability. In \cite{9076290}, an R-DBN-based inverse model is proposed for the first time for modeling a sixth-order and a fourth-order multicoupled cavity filter. The learning time is relatively long, no physical implementation is reported, and only 30 S-parameter samples are used, which raises concerns about generalizability. Recent advances, such as \cite{Karahan2024}’s generalized deep-learning inverse design for complex multiport RF components, enable rapid synthesis of arbitrary-shaped electromagnetic structures with tailored radiative/scattering properties. As demonstrated in \cite{10867510}, a DL-assisted optimization method is introduced to facilitate filter design; however, it still relies on PSO optimization and full-wave EM simulations, and thus is not a purely data-driven approach. 

Among the limited work focused on SIW filters, \cite{9184562} presents a method combining an inverse neural network with a K-impedance inverter model to synthesize a fifth-order H-plane SIW iris filter. While this method avoids conversion errors via de-embedding, thereby avoiding conversion errors associated with equivalent rectangular waveguide models, it is limited to a fixed-order filter, is trained on a relatively small dataset, and does not accommodate a range of SIW topologies or structural variations. Furthermore, it employs conventional neural network architectures, lacking recent advances in deep learning design and training methodologies. Likewise, \cite{9223952} presents an invertible neural networks (INNs) for designing D-band SIW filters. Although this method addresses non-unique solutions, it covers only two design parameters and is trained on a small dataset. Similarly, \cite{Soundarya2023} achieves a 24\%  fractional bandwidth in Ku-band SIW filters with the help of optimization using a feed-forward backpropagation neural network. However, their approach relies on a simple ANN trained on a limited dataset, which constrains modeling accuracy and generalizability. In addition to ANN-based methods,~\cite{10354047} presents a binary optimization approach with notable speedup, yet the limitations in adaptability and generalization remain unaddressed. To better contextualize the novelty of this work, a comparative table summarizing recent SIW inverse design techniques, including their datasets, methods, and key results, is provided in \hyperref[tab:comparison_methods]{Table 1}.

\begin{table}[htbp]
	\begin{minipage}[t]{\textwidth}
	\raggedright
	{\small \textbf{Table 1} \\
		Comparative overview of inverse design methods} \vspace{0.8em}
\end{minipage}
	
	\centering
	\footnotesize
	\setlength{\tabcolsep}{4pt} 
	\renewcommand{\arraystretch}{1.1}
	
	\begin{tabular}{cccccc}
		\toprule
		\textbf{Reference} & \textbf{Method} & \textbf{Dataset Size} & \textbf{Complexity} & \textbf{Generalizability} & \textbf{Runtime Cost} \\
		\midrule
		\cite{9184562}  & ANN & 100 & Low & Moderate & Fast (few ms) \\
		\cite{9223952}  & ANN & 150 & Low & Moderate & Fast (few ms) \\
		\cite{Soundarya2023} & ANN & 386 & Low & Moderate &  Fast (few ms) \\
		\cite{10354047} & Genetic Algorithm & - & Medium & Low & Slow (5–10 s) \\
		\bottomrule
	\end{tabular}
	\label{tab:comparison_methods}
\end{table}

This paper presents the design, fabrication, and experimental validation of multimode SIW filters capable of generating multiple resonances, including both closely spaced and widely separated frequency responses. This outcome is enabled by a novel deep learning framework structured into three progressive stages. This achievement is driven by a novel deep learning framework structured into three progressive stages. First, a conventional Feedforward Inverse Model (FIM) is employed to generate an initial estimate of the design parameters. This model establishes a baseline prediction by mapping S-parameters to the corresponding design parameters. Next, this initial estimate is enhanced using a Hybrid Inverse-Forward Residual Refinement Network (HiFR$^2$-Net), which combines inverse prediction with forward validation and residual correction. By incorporating the forward response within the learning process, this hybrid approach refines the design parameter estimation through iterative residual feedback, improving accuracy and robustness. To further optimize the prediction process, an Iterative Residual Correction Network (IRC-Net) is introduced. This architecture integrates principles from classical iterative refinement algorithms \cite{Stetter1978,KAUCHER1984142} into the deep learning framework. Unlike the previous models, the IRC-Net directly refines the initial design estimate through multiple residual correction cycles, without reprocessing the scattering parameters. This mathematical integration, inspired by IRC methods originally developed for solving Volterra integral equations and linear systems, allows for systematic error reduction and improved convergence. Compared to the baseline FIM and the hybrid HiFR$^2$-Net, our proposed IRC-Net demonstrates superior performance in inverse design tasks, enabling high-precision parameter estimation with minimal reliance on extensive electromagnetic simulations. Experimental validation confirms the effectiveness of this approach, showing low prediction error and strong generalization capabilities across diverse design scenarios.

To provide a comprehensive overview of our methodology and results, the remainder of this paper is organized as follows. Section II presents the design of the multimode resonant SIW structure, detailing its physical layout, parameter definitions, and the results of simulation-based investigations. Section III introduces our deep learning methodology, comprising the preparation of a simulation-driven dataset, development of a feedforward inverse model, and refinement through hybrid inverse-forward and iterative residual correction networks. Section IV discusses the learning behavior and performance evaluation of the proposed models, including statistical analysis and practical validation through simulation and fabrication. Finally, Section V concludes the paper, summarizing key findings and outlining directions for future research.

\section{Design of Multi Mode Resonant SIW Structure}
\label{sec:siw_design}
\subsection{SIW Structure Overview}
SIW technology has emerged as a compelling solution for high-frequency applications, particularly in the Ku-band spectrum. SIW structures emulate the characteristics of conventional rectangular waveguides while offering the advantages of planar circuit integration, such as compactness, low cost, and ease of fabrication using standard PCB processes.

SIW typically consists of a dielectric substrate sandwiched between two metal planes, with rows of metalized via holes forming the sidewalls of the waveguide. These vias confine the electromagnetic fields within the substrate, supporting the propagation of transverse electric (TE) modes. 

The effective width ($W_{eff}$), which governs the cutoff frequency of the structure, can be approximated as follows~\cite{1031925}:

\begin{equation}
	W_{eff}= W - \frac{d^2}{0.95p}
\end{equation}

\noindent{where} $W$ is the width of the structure, $d$ is the diameter of the vias, and $p$ is the spacing between adjacent vias.

To minimize radiation losses and ensure proper wave confinement, the following design rules are typically observed:

\begin{equation}
	p < \frac{\lambda_g}{4} , d < p
\end{equation}

\noindent{here}, $\lambda_g$ is the guided wavelength within the SIW. Adherence to these constraints helps prevent leakage between the vias and maintains the integrity of the waveguide mode. See \cite{1643580} for more details.

Multi-resonant SIW structures are formed by embedding metallic posts within the SIW waveguide. These posts exhibit inductive behavior, and their proper placement enables the creation of multiple resonant frequencies. By carefully selecting the number, size, and positions of the posts, the resonances can be brought closer together, resulting in a bandpass filtering response \cite{article,Salehian2023}. Furthermore, by increasing the frequency spacing between the resonances, it is also possible to design dual-band or multi-band filters, providing more flexibility for various communication and sensing applications \cite{SEDIGHIMARAGHEH2019152885,WU2024154975,10379032}.
The primary degrees of freedom in the design process are the spacing between the posts, the radius of each post, and their specific locations within the structure.

Our main objective in this section is to demonstrate the occurrence of multiple resonances and to analyze how these resonances vary with different structural parameters. Accordingly, design parameters are selected to produce multiple resonances at distinct frequencies. In the case of filter design, where closely spaced resonances are desired, the inductive posts are placed at locations with maximum current density, since they exhibit inductive behavior. Given that the field distribution in SIW structures is analogous to that of rectangular waveguides, these high-current regions can be easily identified, enabling an effective and well-informed post placement strategy.

The following section presents the design of an SIW structure incorporating the aforementioned characteristics. The design aims to achieve a multi-resonant response by exploiting the inductive nature of the posts and carefully tuning their geometrical and positional parameters. Full-wave simulations are employed to analyze how variations in these parameters affect the resonant frequencies, thereby demonstrating the potential of such structures for compact, planar filter implementations.

\subsection{Physical Layout, Parameter Definition and Result of Simulation}

\begin{figure}[htbp]
  \centering
  \includegraphics[width=0.4\textheight]{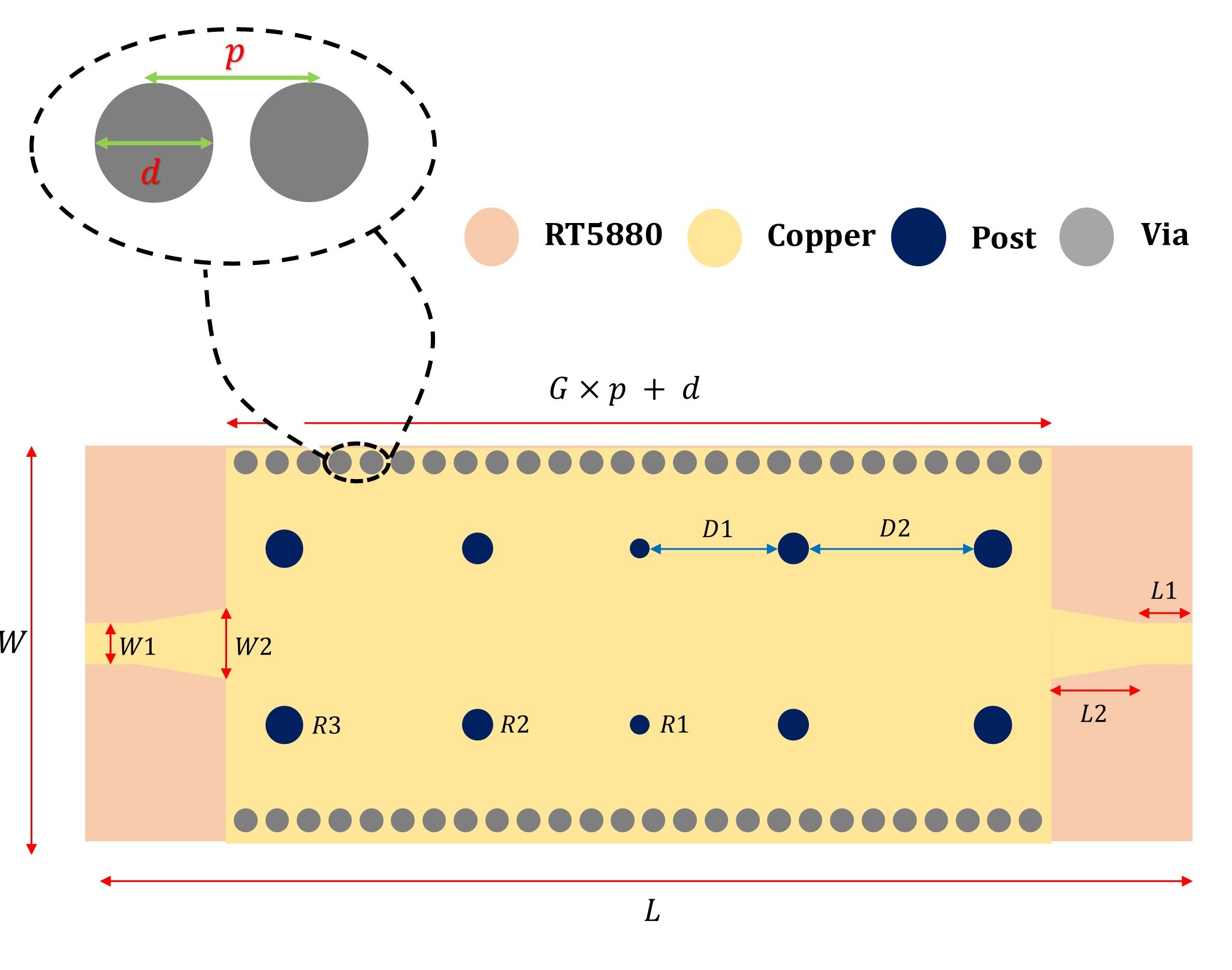}
  \caption{Schematic diagram of the multi-mode SIW structure.}
  \label{fig:siw_structure}
\end{figure}

Figure~\ref{fig:siw_structure} illustrates the schematic of the proposed multi-resonant SIW structure, implemented on an RT5880 substrate with a thickness of 20 mils. The structure incorporates metallic vias forming the sidewalls of the waveguide, along with embedded posts distributed along the propagation path. The geometry of the structure is characterized by several critical parameters: post-to-post spacings ($D_1$, $D_2$), the radius of the coupling vias ($R_1$, $R_2$, $R_3$), and the offset of the posts from the sidewalls. The sidewall via diameter ($d$) and the via pitch ($p$) are fixed to ensure proper operation within the Ku-band.

The spacing between the transition line and the first post can be adjusted using the parameter $G$, which affects the coupling region and is defined as $G \cdot p + d$. Additional geometric parameters include the tapering width $W_2$, matching width $W_1$, tapering length $L_2$, and matching length $L_1$. The overall filter dimensions, denoted by $W$ and $L$, are also indicated in the Figure~\ref{fig:siw_structure}.

As mentioned earlier, the posts act as inductive elements and should therefore be positioned at locations of maximum current within the SIW structure. It is also well established that increasing the post diameter results in higher inductance. In conventional bandpass filter design, these inductive elements are typically arranged in ascending order—i.e., from smaller to larger diameters to achieve the desired frequency response.

In this section, the focus is on designing a multi-mode resonator structure that enables the investigation of how geometric variations influence the frequency shift of each individual resonance. To achieve this, the posts are arranged in descending order of diameter, from larger to smaller, which facilitates differentiation among the resonant modes and allows for observation of their spacing, whether close or wide apart.

\begin{figure}[htbp]
  \centering
  \includegraphics[width=0.4\textheight]{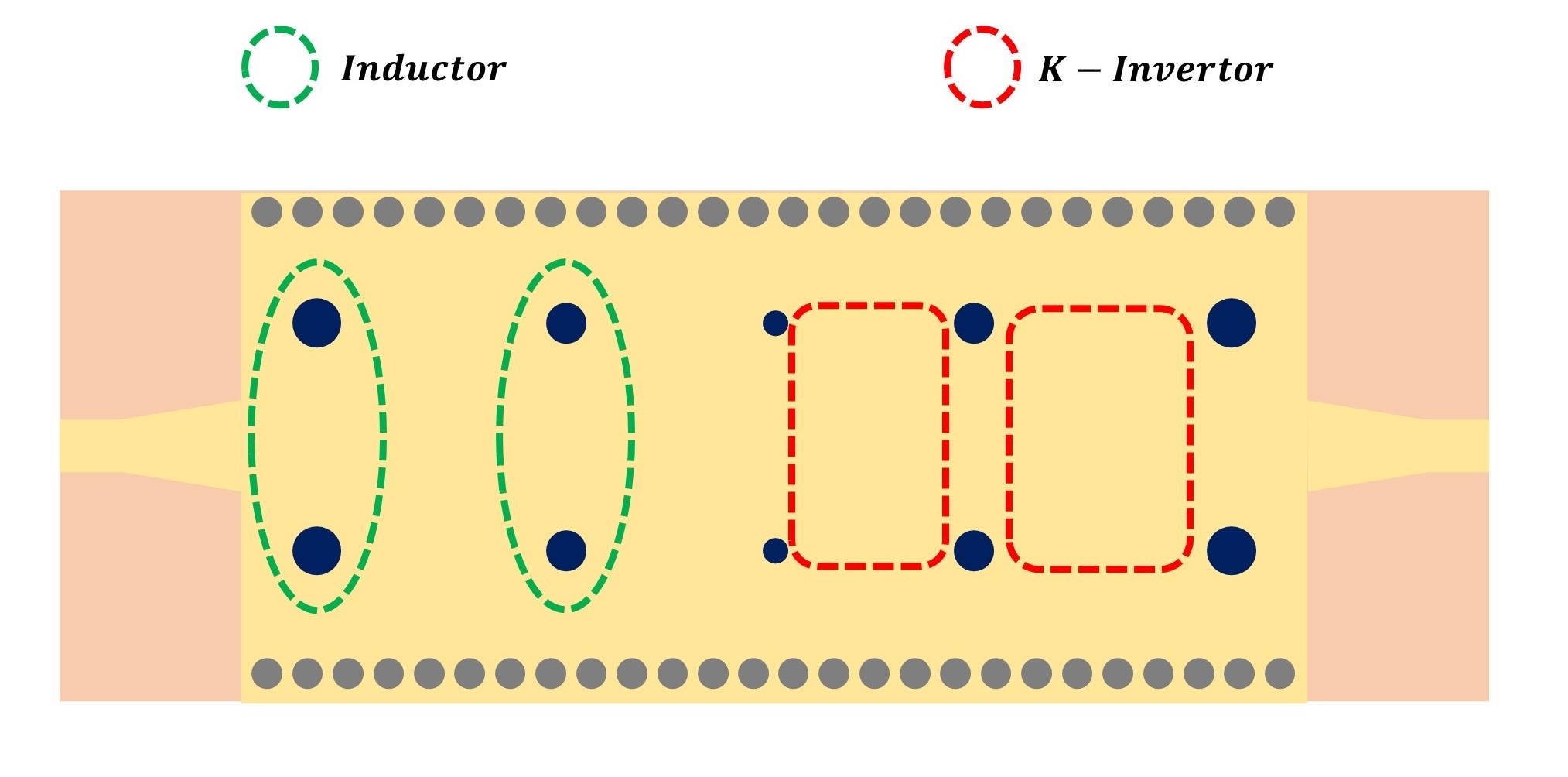}
  \caption{Equivalent circuit of the multi-mode SIW structure.}
  \label{fig:siw_structure3}
\end{figure}

Moreover, since the posts must be placed at the maximum current locations, their spacing is approximately set at $\lambda_g/2$. For further details and more information, see \cite{doi:10.1049/PBEW021E}. The equivalent circuit of the proposed structure, illustrated in Figure~\ref{fig:siw_structure3}, uses $K$-inverters to represent the coupling between adjacent resonators.

This structure is capable of generating four distinct resonant frequencies within the operating band. Moreover, by bringing two of the resonances closer together, it is possible to realize structures with three or even two resonant frequencies, offering more flexibility for various filter design requirements. It is also possible to obtain additional resonances within the band by increasing the number of resonators and the overall length of the structure.
Additionally, multi-section structures can be used to create different impedance profiles, providing more flexibility in tuning the resonant frequencies. This allows the designer to either bring the resonances closer together or space them further apart, depending on the desired filter response. However, the main goal here is to demonstrate that the proposed structure can generate resonances at frequencies specified by the designer. The geometric parameters used to achieve the four in-band resonances are listed in \hyperref[tab:siw_dimensions]{Table 2}. These values were initially chosen based on well-known design guidelines.

\begin{figure}[t!]
  \centering
  \includegraphics[width=0.5\textheight]{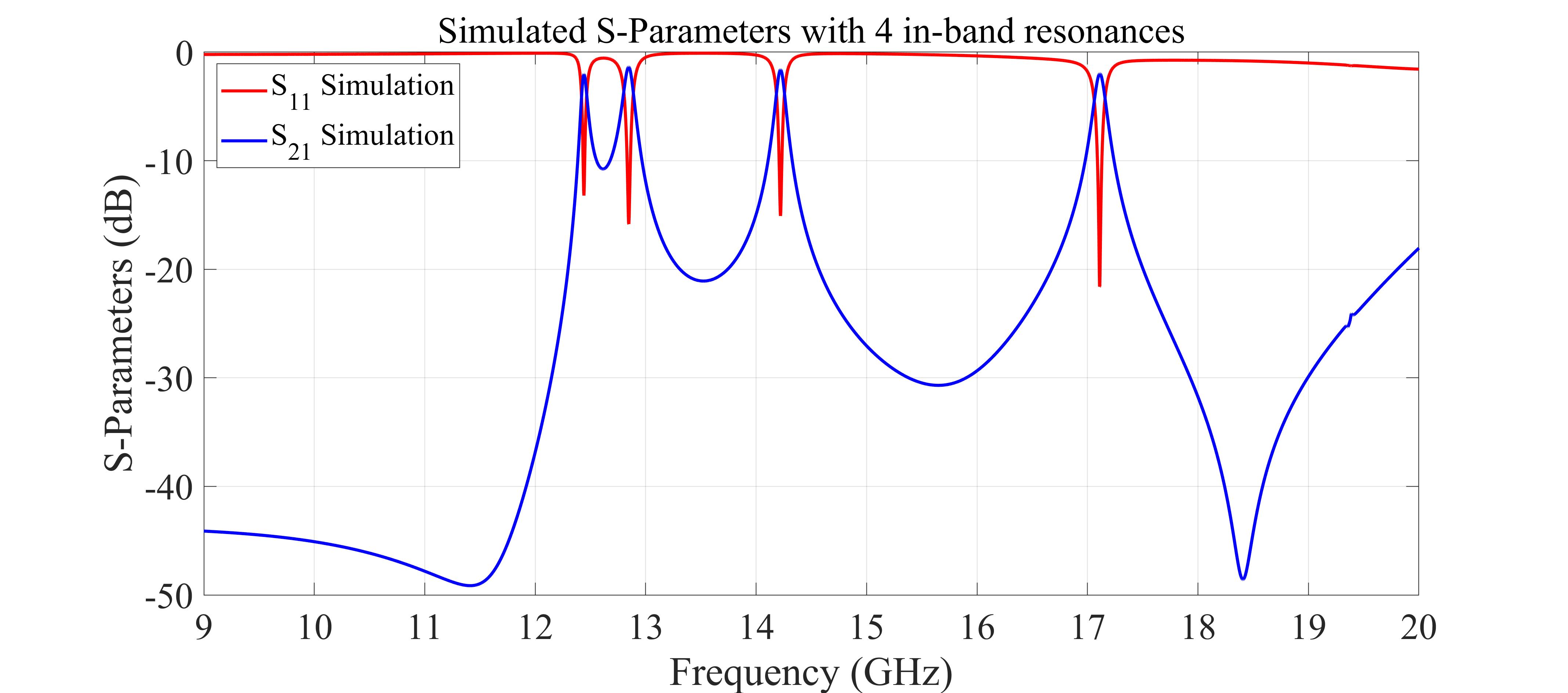}
  \caption{Simulated S parameters ($S_{11}$ and $S_{21}$) of the proposed multi-mode SIW structure with 4 in-band resonances.}
  \label{fig:sparams}
\end{figure}

To evaluate the electromagnetic performance of the proposed filter, full-wave simulations were performed using CST Microwave Studio. Figure~\ref{fig:sparams} presents the simulated S parameters ($S_{11}$ and $S_{21}$) for the four in-band resonances. The reflection coefficient ($S_{11}$) exhibits sharp notches at the desired center frequencies, indicating strong impedance matching. Meanwhile, the transmission coefficient ($S_{21}$) shows a low insertion loss throughout the passband, confirming efficient signal propagation and validating the effectiveness of the designed SIW multimode resonator.

\begin{table}[htbp]
\begin{minipage}[t]{\textwidth}
\raggedright
{\small \textbf{Table 2} \label{tab:siw_dimensions} \\
Geometrical parameters of the multi-mode SIW structure for three in-band resonances.}
\vspace{0.8em}
\end{minipage}

\centering
{\small
\begin{tabular}{ccc}
\toprule
\textbf{Parameter} & \textbf{Description} & \textbf{Value (mm)} \\
\midrule
$W$    & Total width of the SIW filter structure                & 15   \\
$W_1$  & Width of the input/output taper section                & 1.5  \\
$W_2$  & Width of the internal SIW waveguide section            & 2    \\
$L$    & Length of the entire SIW filter                        & 42.2 \\
$L_1$  & Length of the input taper section                      & 1    \\
$L_2$  & Length of the output taper section                     & 3    \\
$R_1$  & Radius of the first coupling via                       & 0.2  \\
$R_2$  & Radius of the second coupling via                      & 0.4  \\
$R_3$  & Radius of the third coupling via                       & 0.8  \\
$D_1$  & Distance between the first and second posts            & 5.5  \\
$D_2$  & Distance between the second and third posts            & 8    \\
$d$    & Diameter of the sidewall vias                          & 0.8  \\
$p$    & Spacing between sidewall vias                          & 1.3  \\
$G$    & Scaling factor                                         & 26   \\
\bottomrule
\end{tabular}
}

\end{table}

\subsection{Investigation of the Effect of Sweeping Design Parameters}
\begin{figure}[t!]
  \centering

  \begin{subfigure}[b]{0.48\linewidth}
    \centering
    \includegraphics[width=\linewidth]{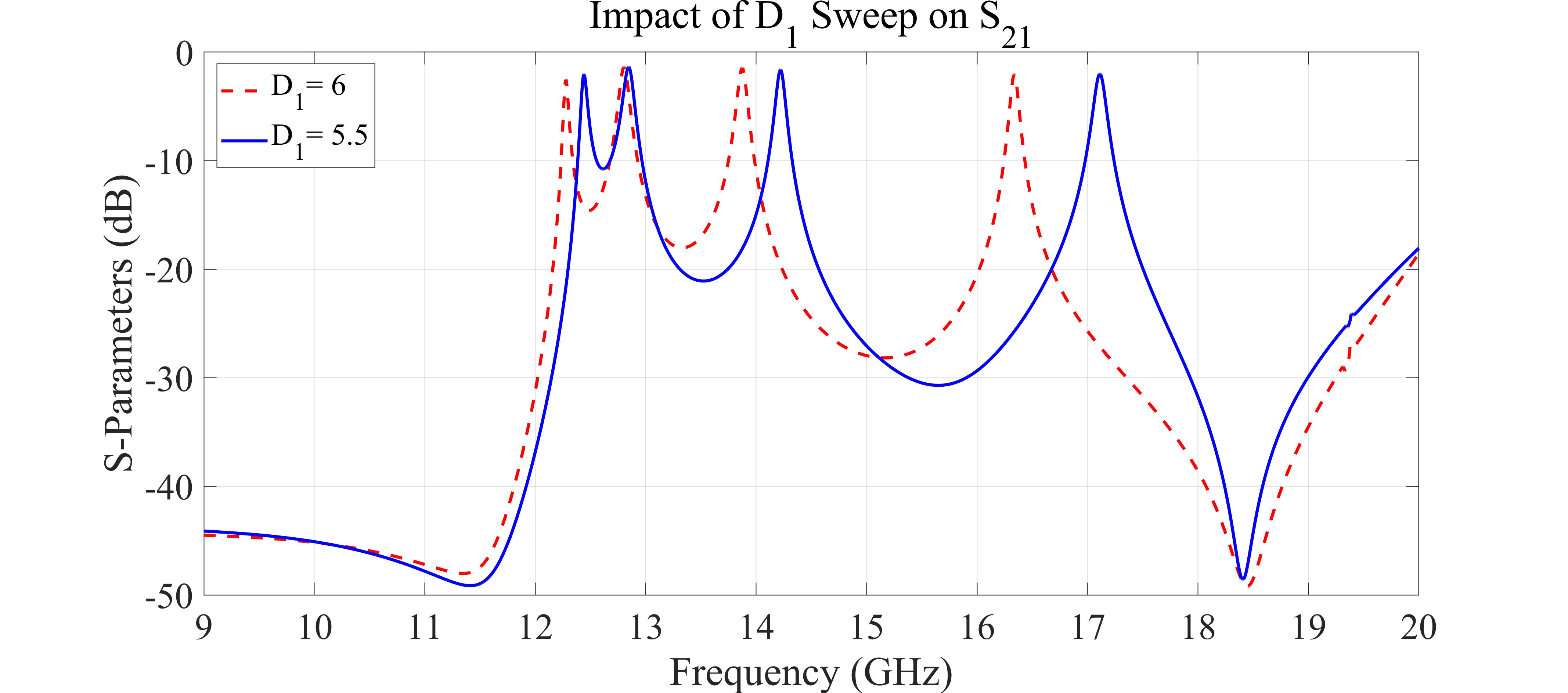}
    \caption{Effect of varying via diameter $D_1$.}
    \label{fig:d1_sweep}
  \end{subfigure}
  \hfill
  \begin{subfigure}[b]{0.48\linewidth}
    \centering
    \includegraphics[width=\linewidth]{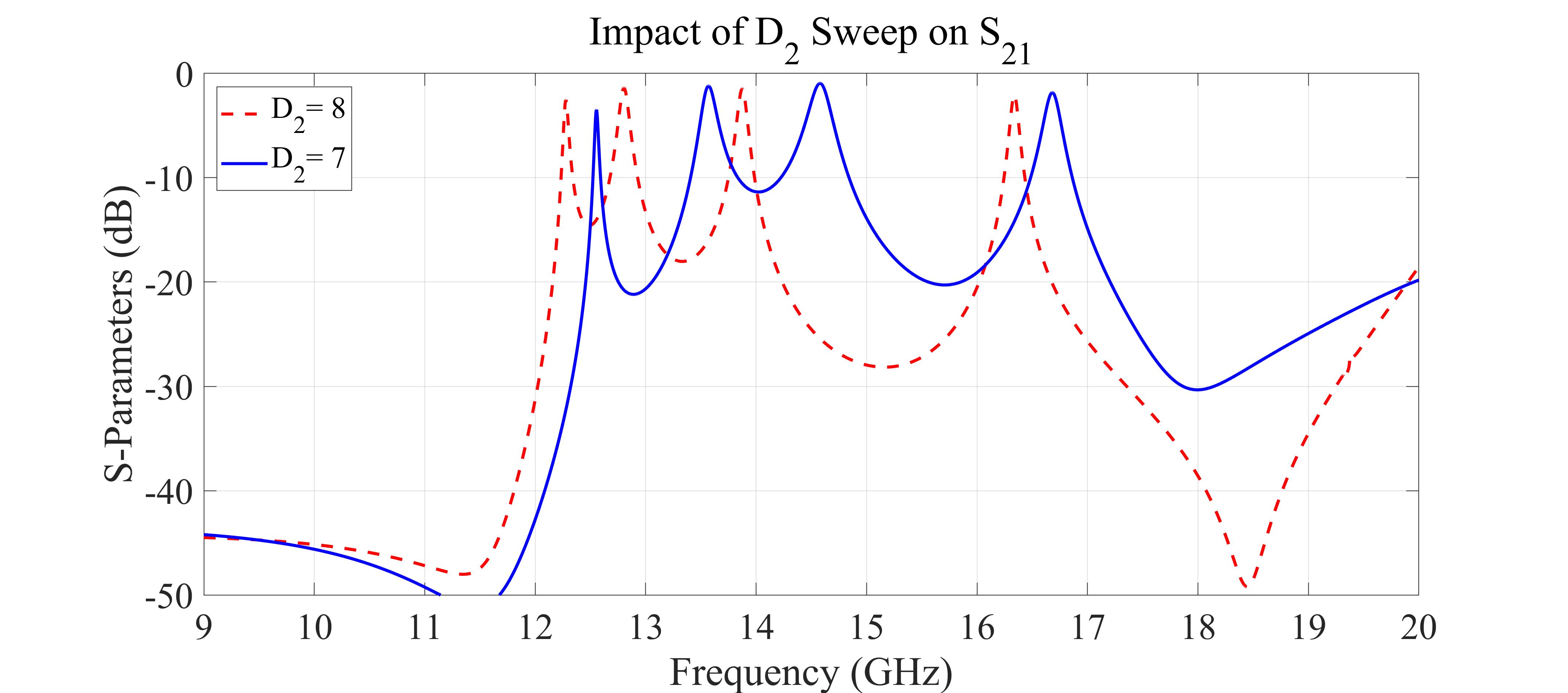}
    \caption{Effect of varying via diameter $D_2$.}
    \label{fig:d2_sweep}
  \end{subfigure}

  \caption{Effect of varying via diameters $D_1$ and $D_2$ on the S-parameters of the multi-mode SIW structure.}
  \label{fig:d_sweep_combined}
\end{figure}

To further understand the behavior of the proposed SIW multi-mode structure, a parametric analysis was performed by sweeping critical geometrical parameters and observing their impact on the filter’s electromagnetic response. Figures~\ref{fig:d1_sweep} and~\ref{fig:d2_sweep} illustrate the effect of varying the diameters $D_1$ and $D_2$, respectively, while keeping other parameters fixed.

The sweep results indicate that both \( D_1 \) and \( D_2 \) have a significant impact on the center frequencies and bandwidths of the resonant modes. Specifically, increasing \( D_1 \) shifts the resonant frequencies to lower values and causes them to move closer together. Similarly, increasing \( D_2 \) also shifts the resonances to lower frequencies; however, it leads to an increased spacing between the resonant modes, resulting in the resonances moving farther apart.

Variations in distances \( D_1 \) and \( D_2 \) alter the coupling strength and, consequently, the characteristics of the \( K \)-inverters. As a result, these changes are expected to affect the behavior of the S-parameters.

Although the parameter sweeps provided useful insights into the roles of \( D_1 \) and \( D_2 \), they also underline a key limitation: as the number of design parameters increases, it becomes significantly more challenging to analyze the effects of each parameter individually. The interdependencies among parameters can lead to complex and sometimes unpredictable behavior, making it difficult to rely solely on parameter sweeps for accurate design optimization. In this study, six key parameters are considered in the design process: \( D_1 \), \( D_2 \), \( R_1 \), \( R_2 \), \( R_3 \), and \( G \). Each of these parameters can significantly influence the resonant frequencies, bandwidth, and coupling strength.

To address the inherent complexity of this problem, deep learning is leveraged to model the forward and inverse relationships between SIW geometries and their corresponding electromagnetic characteristics. By generating a comprehensive simulation-based dataset and training a fully connected neural network, our approach aims to:

\begin{itemize}
  \item Rapidly predict the design parameters that produce a broad range of customized responses in $S_{11}$ and $S_{21}$.
  \item Accurately perform inverse design, mapping desired frequency characteristics to optimal physical geometries.
  \item Significantly reduce the computational time and resource requirements compared to traditional full-wave electromagnetic simulations.
\end{itemize}

This data-driven approach is particularly suited to multi-resonant SIW structures, where multiple tightly coupled parameters must be tuned to meet stringent filter specifications. The next section details the proposed deep learning framework and dataset generation process.

\section{Deep Learning Methodology}
\label{sec:dl_method}

This section outlines the deep learning framework developed for the inverse design of multimode SIW filters. The methodology is structured into four key stages, each building upon the previous to progressively enhance prediction accuracy and robustness. The process begins with Dataset Preparation, which details the generation and preprocessing of input-output pairs—comprising S-parameters and design parameters—for neural network training. Next, the Feedforward Inverse Model (FIM)is introduced as the foundational network that estimates the filter design based on measured S-parameters. Building on this foundation, the Hybrid Inverse-Forward Residual Refinement Network (HiFR\textsuperscript{2}-Net) is presented, incorporating both inverse and forward mappings to refine predictions through residual correction. Finally, the Iterative Residual Correction Network (IRC-Net) is described, employing successive correction loops to iteratively adjust the design estimate, thereby aligning it more closely with the true design parameters.

This progressive approach demonstrates how deep learning architectures can be adapted and enhanced to solve complex inverse design problems with increasing levels of precision and robustness.

\subsection{Dataset Preparation}
The dataset used in this study comprises \textbf{8,721 samples}, each corresponding to a unique configuration of an SIW multimode resonator. Each configuration is defined by six geometric parameters: \(D_1\), \(D_2\), \(R_1\), \(R_2\), \(R_3\), and \(G\). These parameters were generated through structured nested loops, constrained by physical and electromagnetic design rules to ensure valid configurations.

Two constraints were applied during data generation to guarantee physically meaningful and practical designs:
\[
R_3 \geq R_2 \geq R_1, \quad \text{and} \quad \frac{\left(R_1 + 2R_2 + D_1 + D_2 + 2R_3\right) \times 2 - 0.4}{1.3} < G.
\]

The first constraint ensures proper multi-mode resonator construction, while the second enforces adequate spacing between resonators to prevent overlapping via structures.

Sampling ranges for the parameters were defined as follows:
\begin{itemize}
    \item \(D_1, D_2 \in [4\,\mathrm{mm}, 10\,\mathrm{mm}]\) with a step size of 0.5\,mm,
    \item \(R_1, R_2, R_3 \in [0.2\,\mathrm{mm}, 1.0\,\mathrm{mm}]\) with a step size of 0.2\,mm,
    \item \(G \in [26\,\mathrm{mm}, 36\,\mathrm{mm}]\).
\end{itemize}

For each valid configuration, full-wave electromagnetic simulations were performed using CST Studio Suite to extract the S-parameter responses. The magnitude responses of \(S_{11}\) and \(S_{21}\) were sampled across the Ku-band frequency range of 9–20\,GHz at 1,001 frequency points, resulting in a total of 2,002 input features per sample.

The entire dataset was generated over approximately 92 hours on a workstation equipped with a 13th Gen Intel(R) Core(TM) i5-13600K CPU running at 3.5 GHz and 128 GB of RAM, highlighting the significant computational cost of conventional simulation methods.

Dataset preparation involved the following steps:
\begin{enumerate}
    \item Magnitude responses of \(S_{11}\) and \(S_{21}\) were loaded from \texttt{.mat} files and concatenated to form the input matrix \(\mathbf{X} \in \mathbb{R}^{8721 \times 2002}\).
    \item Corresponding six-dimensional geometric parameters were collected and normalized, forming the output matrix \(\mathbf{Y} \in \mathbb{R}^{8721 \times 6}\).
    \item All features and targets were standardized to zero mean and unit variance.
    \item The dataset was split into 80\% for training and 20\% for testing, using a fixed random seed (42) to guarantee reproducibility.
\end{enumerate}

This prepared dataset was utilized for training both the primary deep neural network and the subsequent iterative residual correction models detailed in the following sections. Dataset preprocessing and model training were conducted on the Google Colab platform, where data preparation scripts leveraged Colab's computational resources for efficient processing, and GPU acceleration significantly expedited model convergence on the large-scale dataset.

\subsection{Feedforward Inverse Model}

The Feedforward Inverse Model (FIM) serves as a traditional baseline for the inverse design problem, providing a straightforward and effective approach to estimating the geometric design parameters. This model has been widely used in various engineering fields due to its simplicity and practical utility. By leveraging a deep fully connected neural network, FIM maps high-dimensional electromagnetic scattering data directly to the corresponding low-dimensional design parameters. The key advantage of FIM lies in its ability to offer a first-order approximation of the inverse mapping, which is typically ill-posed and highly nonlinear in many real-world problems.

The primary objective of FIM is to predict the geometric design parameters \(\mathbf{P}_0\), which serve as an initial estimate of the true design vector \(\mathbf{D}_{\text{true}}\). This is achieved solely based on the S-parameter data, without requiring detailed prior knowledge of the geometry. However, while effective in many cases, the simplicity of FIM can limit its performance, particularly when dealing with complex or highly nonlinear mappings. To address these challenges, FIM’s universal approximation capability, as a deep neural network, allows it to capture intricate dependencies across a wide range of SIW geometries.

The model takes as input a 2,002-dimensional vector, consisting of 1,001 frequency samples for both \(S_{11}\) and \(S_{21}\). It is structured with seven fully connected layers that progressively decrease in size, facilitating efficient feature extraction and compression. The ReLU activation function is employed throughout to introduce nonlinearity, while dropout regularization with a fixed rate of 10\% is applied after each hidden layer to prevent overfitting. The model is trained using the mean squared error (MSE) loss function and optimized via the Adam optimizer. To prevent overtraining and ensure the model generalizes well to unseen data, early stopping is also implemented. A detailed summary of the model architecture is presented in \hyperref[tab:FIM_model]{Table 3}.

This FIM provides the coarse inverse prediction \(\mathbf{P}_0\), which forms the basis for subsequent refinement in the overall framework. Although \(\mathbf{P}_0\) may not precisely match \(\mathbf{D}_{\text{true}}\), it captures the dominant structural trends required for meaningful correction. In the next stage, this estimate forms the basis for further refinement in the overall framework, which is detailed in the subsequent section, where the residual correction mechanism is introduced.

\begin{table}[htbp]
\begin{minipage}[t]{\textwidth}
\raggedright
{\small \textbf{Table 3} \\
Architecture of the Feedforward Inverse Network.} \vspace{0.8em}
\end{minipage}

\centering
{\small
\begin{tabular}{cccc}
\toprule
\textbf{Layer} & \textbf{Size / Type} & \textbf{Activation} & \textbf{Description} \\
\midrule
Input     & 2002 (Dense)  & ReLU & Concatenated S-parameter vector \\
Hidden 1  & 1500 (Dense)  & ReLU & Initial fully connected layer \\
Hidden 2  & 1000 (Dense)  & ReLU & Intermediate layer \\
Dropout 1 & 10\%          & --   & Regularization \\
Hidden 3  & 500 (Dense)   & ReLU & Intermediate layer \\
Dropout 2 & 10\%          & --   & Regularization \\
Hidden 4  & 250 (Dense)   & ReLU & Intermediate layer \\
Dropout 3 & 10\%          & --   & Regularization \\
Hidden 5  & 125 (Dense)   & ReLU & Intermediate layer \\
Dropout 4 & 10\%          & --   & Regularization \\
Hidden 6  & 64 (Dense)    & ReLU & Intermediate layer \\
Dropout 5 & 10\%          & --   & Regularization \\
Hidden 7  & 32 (Dense)    & ReLU & Final hidden layer \\
Output    & 6 (Dense)     & ReLU & Estimated design parameters $\mathbf{P}_0$ \\
\bottomrule
\end{tabular}
}
\label{tab:FIM_model}
\end{table}

\subsection{Hybrid Inverse-Forward Residual Refinement Network}

Although the FIM offers a straightforward and computationally efficient approach for estimating geometric parameters from S parameters, its simplicity may hinder accurate modeling of complex inverse dynamics, particularly in ill-posed or highly nonlinear scenarios.

\begin{figure}[t!]
    \centering
    \includegraphics[width=0.8\linewidth]{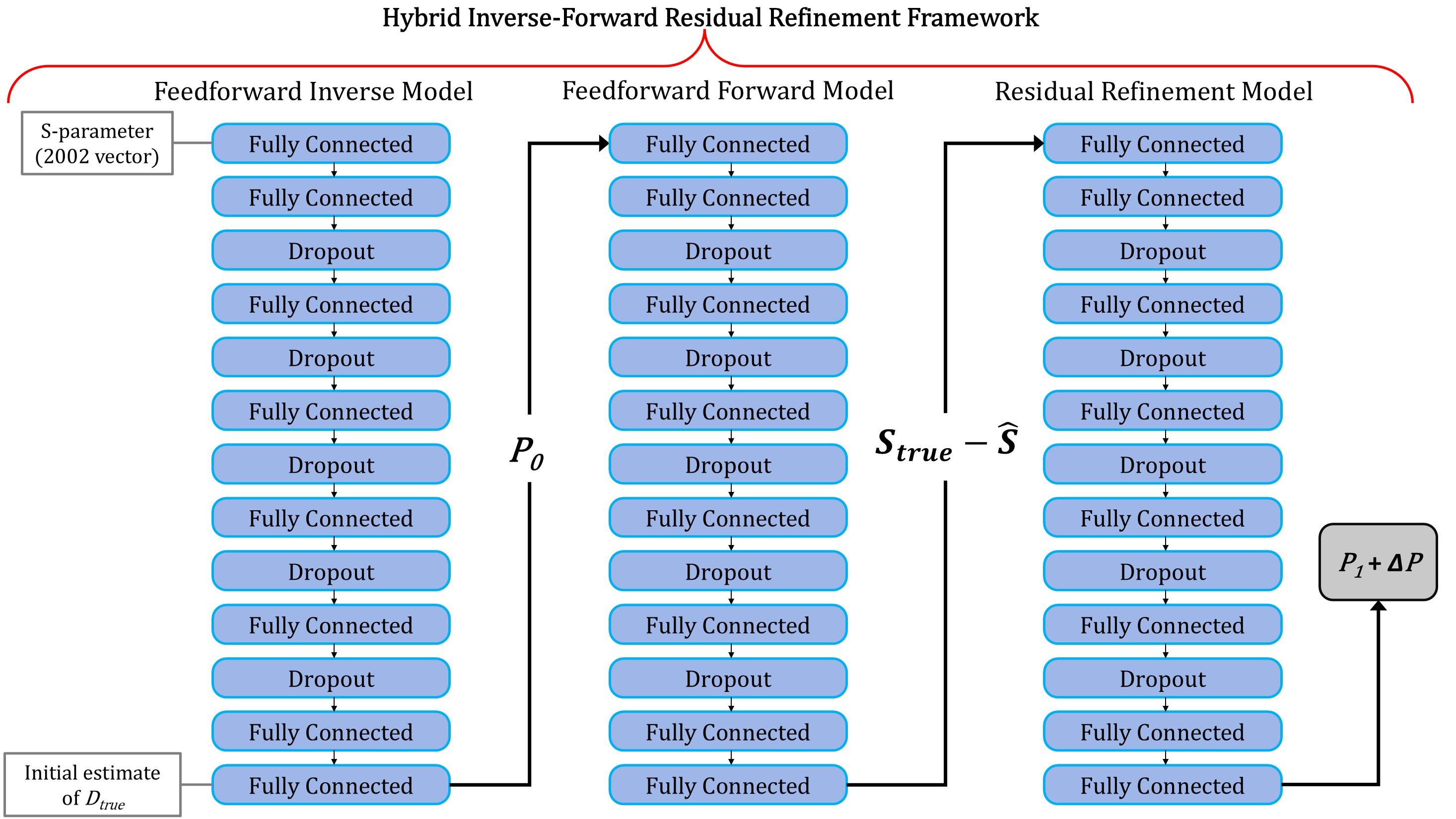}
    \caption{Schematic overview of the Hybrid Inverse-Forward Residual Refinement Network.}
    \label{fig:HiFR2Net}
\end{figure}

To enhance the prediction accuracy and consistency of inverse modeling, the Hybrid Inverse-Forward Residual Refinement Network (HiFR\textsuperscript{2}-Net) is introduced, as schematically depicted in Figure~\ref{fig:HiFR2Net}. This cascaded learning framework integrates both inverse and forward learning stages, along with a residual refinement mechanism, to iteratively improve the fidelity of the design parameter estimation.

The HiFR\textsuperscript{2}-Net consists of the following three stages:

\begin{itemize}
    \item \textbf{Stage 1 – Feedforward Inverse Model:} As previously detailed, the FIM serves as a baseline inverse predictor that generates an initial estimate \(\mathbf{P}_0\) of the normalized geometric design parameters from the input S parameters.
    
    \item \textbf{Stage 2 – Feedforward Forward Model:} The initial prediction \(\mathbf{P}_0\) is fed into a forward model trained to reconstruct the original S-parameter vector \(\hat{\mathbf{S}}\). This stage enforces forward consistency and enables indirect validation of the initial estimate through reconstruction error.
    
    \item \textbf{Stage 3 – Residual Refinement Model:} The residual vector, computed as the difference between the true and reconstructed S parameters (\(\mathbf{S}_{\text{true}} - \hat{\mathbf{S}}\)), is used as input to a refinement model. This model predicts a correction term \(\Delta\mathbf{P}\), which is added to the initial estimate \(\mathbf{P}_0\) to improve alignment with the true design parameters.
\end{itemize}

The architecture of the FFM is summarized in \hyperref[tab:ffm_model]{Table 4}. It consists of multiple fully connected layers that progressively expand the input representation to reconstruct the high-dimensional S-parameter vector.

The RRM is identical to that of the FIM presented in \hyperref[tab:FIM_model]{Table 3}, with the exception of its input and output: the RRM takes residual S parameters ($\mathbf{S}_{\text{true}} - \hat{\mathbf{S}}$) as input and predicts a correction to be added to the original parameter estimate ($\Delta\mathbf{P}$).

The integration of forward reconstruction and residual learning enables the system to better capture hidden dependencies and correct systematic estimation errors. The next subsection extends this concept by introducing an iterative correction mechanism that refines predictions through multiple passes.

\begin{table}[htbp]
\begin{minipage}[t]{\textwidth}
\raggedright
{\small \textbf{Table 4} \\
Architecture of the Feedforward Forward Network.}
\vspace{0.8em}
\end{minipage}

\centering
{\small
\begin{tabular}{cccc}
\toprule
\textbf{Layer} & \textbf{Size / Type} & \textbf{Activation} & \textbf{Description} \\
\midrule
Input     & 6 (Dense)       & ReLU & Estimated parameters $\mathbf{P}_0$ from FIM \\
Hidden 1  & 32 (Dense)      & ReLU & Initial expansion layer \\
Dropout 1 & 10\%            & --   & Regularization \\
Hidden 2  & 64 (Dense)      & ReLU & Intermediate representation \\
Dropout 2 & 10\%            & --   & Regularization \\
Hidden 3  & 128 (Dense)     & ReLU & Intermediate representation \\
Dropout 3 & 10\%            & --   & Regularization \\
Hidden 4  & 256 (Dense)     & ReLU & Feature expansion \\
Dropout 4 & 10\%            & --   & Regularization \\
Hidden 5  & 512 (Dense)     & ReLU & Deep representation \\
Dropout 5 & 10\%            & --   & Regularization \\
Hidden 6  & 1024 (Dense)    & ReLU & High-capacity layer \\
Output    & 2002 (Dense)    & ReLU & Reconstructed output vector $\hat{\mathbf{S}}$ \\
\bottomrule
\end{tabular}
}
\label{tab:ffm_model}
\end{table}

\subsection{Iterative Residual Correction Network}

The iterative residual correction Network (IRC-Net) plays a central role in enhancing the accuracy of the inverse design process. While traditional deep learning-based inverse models can yield useful initial estimates, they often fall short in capturing the full complexity of the design space due to the ill-posed and nonlinear nature of the problem. 

While the concept of iterative correction has been explored in other domains, its structured integration with inverse-forward consistency and residual learning in the context of SIW multimode resonators represents a tailored and effective solution for high-fidelity inverse modeling.

The core idea behind this approach is to compute and correct the residuals—i.e., the differences between the predicted parameters and the true design parameters \(\mathbf{D}_{\text{true}}\)—at each iteration. This correction process continues until the predictions converge to a more accurate representation of the true design. Importantly, each iteration uses the same fixed baseline output \(\mathbf{P}_0\) from the FIM, ensuring that the initial design context remains consistent throughout the iterative process. The residuals are updated iteratively, allowing the model to gradually hone in on the correct design parameters.

This approach is particularly powerful because it enables a more detailed correction of the initial prediction by focusing on smaller, residual discrepancies that may not be captured in a single forward pass through the network. Furthermore, the modular design of the residual correction network allows for a highly flexible and interpretable architecture, making it suitable for a wide range of inverse design problems in electromagnetics and beyond.

Each iteration \(i\) involves the following steps:
1. The residual between the current prediction \(\mathbf{P}_{i-1}\) and the true design parameters \(\mathbf{D}_{\text{true}}\) is computed.
2. This residual is then modeled by a residual network, which outputs a 6-dimensional correction vector \(\Delta \mathbf{P}_i\).
3. The residual correction is added to the previous prediction:
   \[
   \mathbf{P}_i = \mathbf{P}_{i-1} + \Delta \mathbf{P}_i
   \]
4. The updated prediction $\mathbf{P}_i$ becomes the input for the next correction cycle, with the process repeating for a fixed number of iterations or until convergence.

This iterative process is repeated for \(T = 5\) iterations, and at the end of this loop, the final refined prediction \(\mathbf{P}_T\) represents a much closer approximation to the true design parameters. The loop’s iterative nature enables the model to address subtle nonlinear discrepancies in the design parameters, significantly enhancing the model's generalization and accuracy.

The architecture of the IRC-Net used in each iteration is summarized in \hyperref[tab:residual_model]{Table 5}. It is a compact feedforward network consisting of two hidden layers with LeakyReLU activations, followed by a linear output layer that produces a 6-dimensional correction vector \(\Delta \mathbf{P}_i\). This correction is then added to the previous prediction \(\mathbf{P}_{i-1}\), progressively refining the model’s estimate of the true design parameters. The use of a lightweight and consistent architecture across iterations ensures both efficiency and effective residual modeling.

\begin{figure}[t!]
    \centering
    \includegraphics[width=0.8\textwidth]{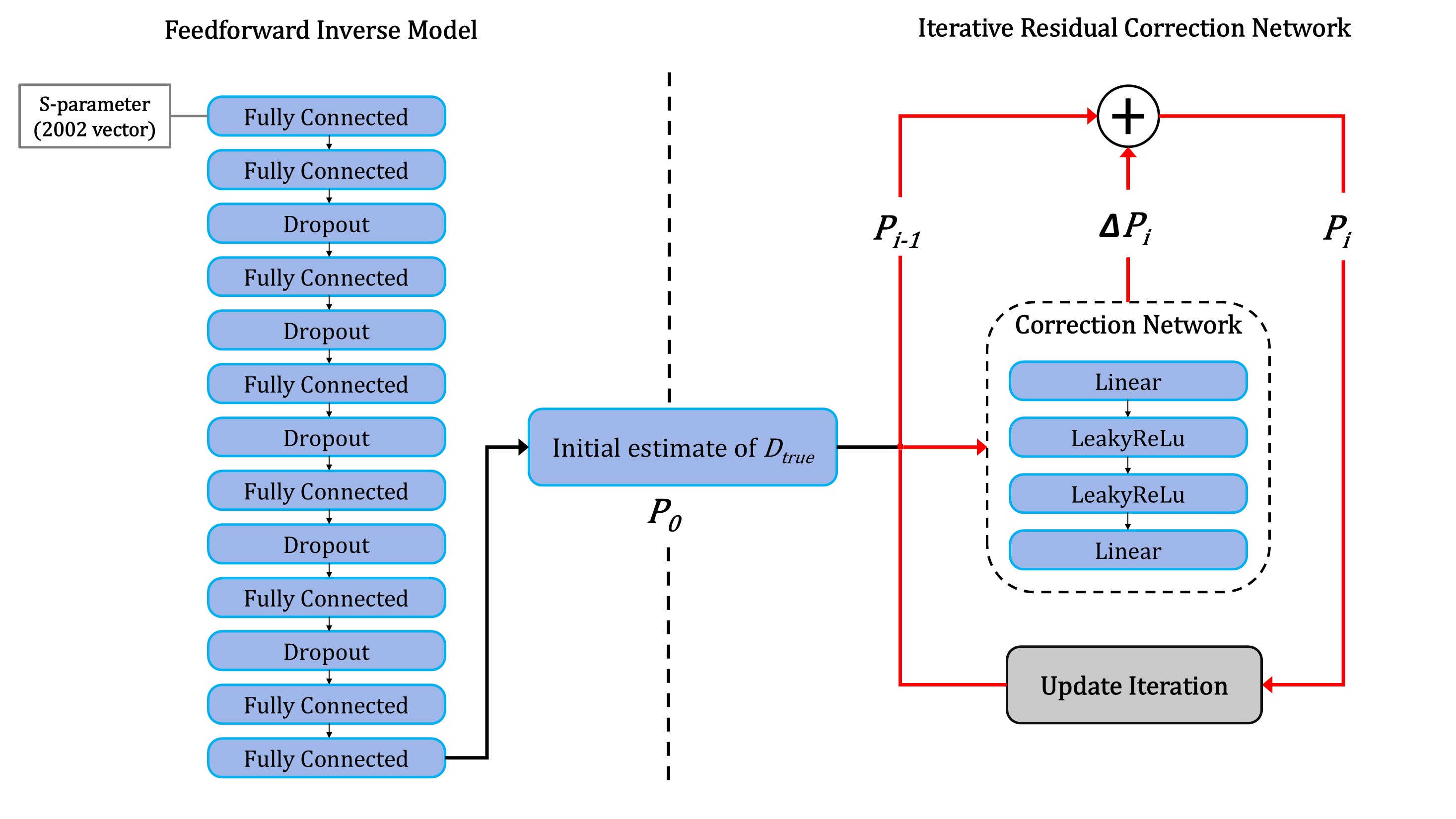}
    \caption{Schematic of the proposed two-stage architecture: a feedforward inverse model followed by iterative residual correction loops.}
    \label{fig:irc_architecture}
\end{figure}

A schematic representation of the proposed two-stage architecture, consisting of the FIM and IRC-Net, is shown in Figure~\ref{fig:irc_architecture}. This diagram visually captures the flow of information and the key iterative refinement process at the heart of the architecture.

This modular architecture provides both flexibility and interpretability. The FIM captures the dominant inverse mapping, while the residual correction loop functions as an adaptive refinement mechanism to address subtle nonlinear discrepancies—thereby improving generalization and robustness in the inverse modeling of complex electromagnetic structures.

\begin{table}[htbp]
\begin{minipage}[t]{\textwidth}
\raggedright
{\small \textbf{Table 5} \label{tab:residual_model} \\ 
Architecture of Residual Correction Network used in each iteration.}
\vspace{0.8em}
\end{minipage}

\centering
{\small
\begin{tabular}{cccc}
\toprule
\textbf{Layer} & \textbf{Size / Type} & \textbf{Activation} & \textbf{Description} \\
\midrule
Input    & 6 (Dense)    & Linear     & Fixed baseline output $\mathbf{P}_0$ \\
Hidden 1 & 64 (Dense)   & LeakyReLU  & Residual mapping layer \\
Hidden 2 & 64 (Dense)   & LeakyReLU  & Deep residual correction \\
Output   & 6 (Dense)    & Linear     & Predicted correction $\Delta \mathbf{P}_i$ \\
\bottomrule
\end{tabular}
}
\end{table}

\section{Results and Discussion}
\label{sec:results_discussion}
This section presents a comprehensive evaluation of the proposed deep learning framework for inverse modeling of SIW multimode resonators. The objective is to analyze the effectiveness, generalization, and practical viability of the models developed in this study. The evaluation includes both statistical assessments and full-wave simulation-based validations.

The section is organized into three main parts: (i) analysis of model learning behavior, focusing on the convergence and predictive capabilities of the FIM, HiFR\textsuperscript{2}-Net, and the IRC-Net; (ii) statistical comparison of model performance using standard metrics; and (iii) practical validation through electromagnetic simulations and application-level implementations. This structure ensures a well-rounded assessment of both theoretical and practical aspects of the proposed methodology.

In order to support the practical viability of the proposed models, we also report the computational cost and training time of each approach. All models were trained on Google Colab using an NVIDIA Tesla T4 GPU, which provides 16~GB of GDDR6 memory and is optimized for both training and inference workloads. The baseline models (FIM and HiFR\textsuperscript{2}-Net) were trained with a batch size of \textbf{128} using early stopping, while the iterative correction stage in IRC-Net employed a batch size of \textbf{32} for \textbf{100} epochs. The total training durations were approximately \textbf{161 seconds} for FIM, \textbf{477 seconds} for HiFR\textsuperscript{2}-Net, and \textbf{867 seconds} for IRC-Net (including 160 seconds for the initial FIM and 707 seconds for 5 iterations of residual correction). The prediction time per sample for all models is less than one millisecond, confirming the suitability of the proposed framework for real-time applications.

\subsection{Model Learning Behavior}

This subsection investigates the learning behavior of the three main components of the proposed framework: the FIM, the HiFR\textsuperscript{2}-Net, and IRC-Net. The effectiveness of these models is assessed based on their ability to minimize the loss function—particularly the Mean Squared Error (MSE) and Mean Absolute Error (MAE)—during training, as well as their generalization capabilities across a set of test cases

\subsubsection{Learning Behavior of the FIM}
\label{subsubsec:Learning Behavior of the FIM}
\begin{figure}[htbp]
    \centering
    \includegraphics[width=0.8\textwidth]{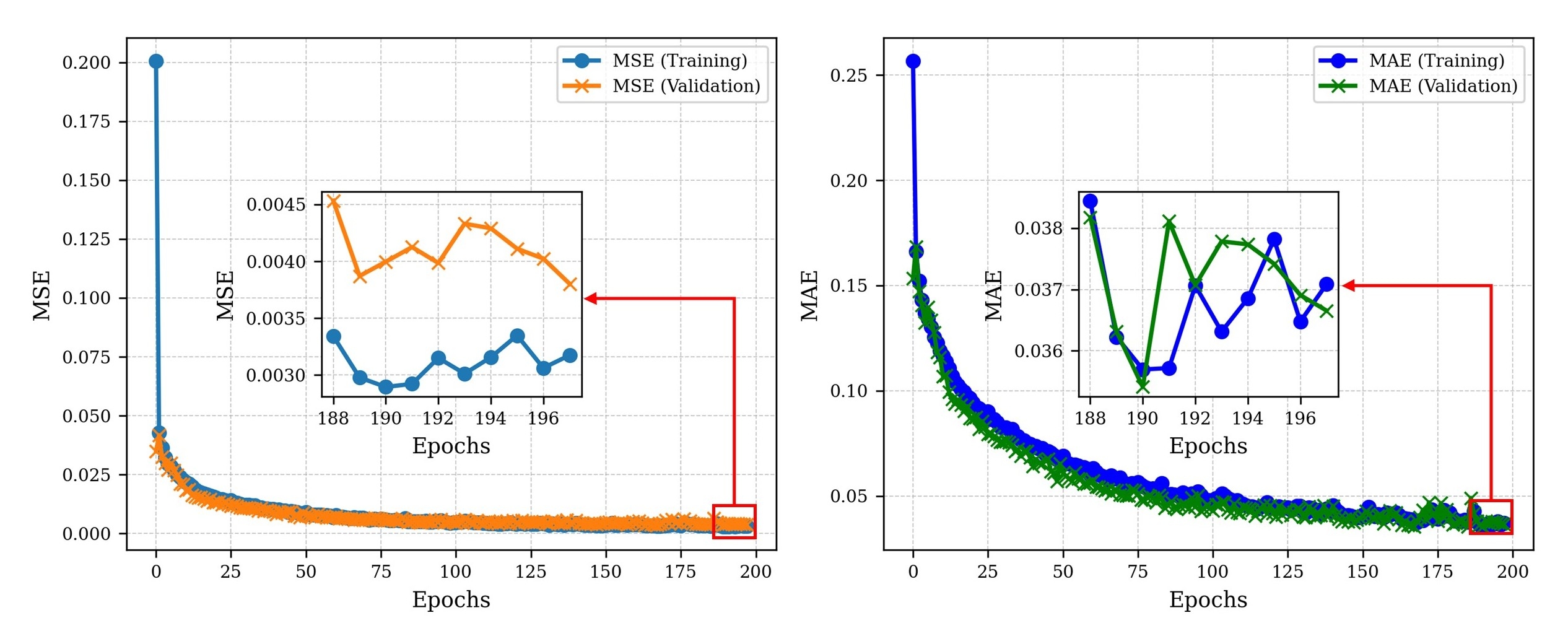}
    \caption{Learning curve for the FIM, showing the convergence of the training and validation losses.}
    \label{fig:fim_learning}
\end{figure}

The learning performance of the FIM is assessed by examining its ability to produce an initial estimate of the design parameters, denoted as \(\mathbf{P}_0\). The corresponding learning curves, depicted in Figure~\ref{fig:fim_learning}, show the evolution of both training and validation errors in terms of MSE and MAE over 200 epochs.

As illustrated in the left panel of Figure~\ref{fig:fim_learning}, the MSE for both the training and validation sets rapidly decreases during the initial stages of training and continues to decline gradually until convergence. Notably, the final training and validation MSE values settle around \(0.0032\) and \(0.0041\), respectively, indicating that the model is capable of learning a reasonable initial mapping from response space to design parameters. A zoomed-in subplot highlights the stability of the learning process in the final epochs.

Similarly, the right panel presents the MAE learning curves, which follow a consistent downward trend. The training MAE reaches approximately \(0.0365\), while the validation MAE stabilizes around \(0.0372\) by the end of training. The tight gap between training and validation errors in both MSE and MAE indicates that the FIM does not suffer from overfitting and generalizes reasonably well within its single-pass prediction capacity.

Overall, the FIM demonstrates efficient and stable convergence with promising results in both training and validation phases. However, despite its effectiveness in providing an initial estimate, its performance can be further enhanced by employing an iterative refinement approach such as the IRC-Net, which incrementally reduces prediction errors and improves model accuracy.

\subsubsection[Learning Behavior of the HiFR2-Net]{Learning Behavior of the HiFR\textsuperscript{2}-Net}

\begin{figure}[t!]
    \centering
    \begin{subfigure}[b]{0.8\linewidth}
        \centering
        \includegraphics[width=\linewidth]{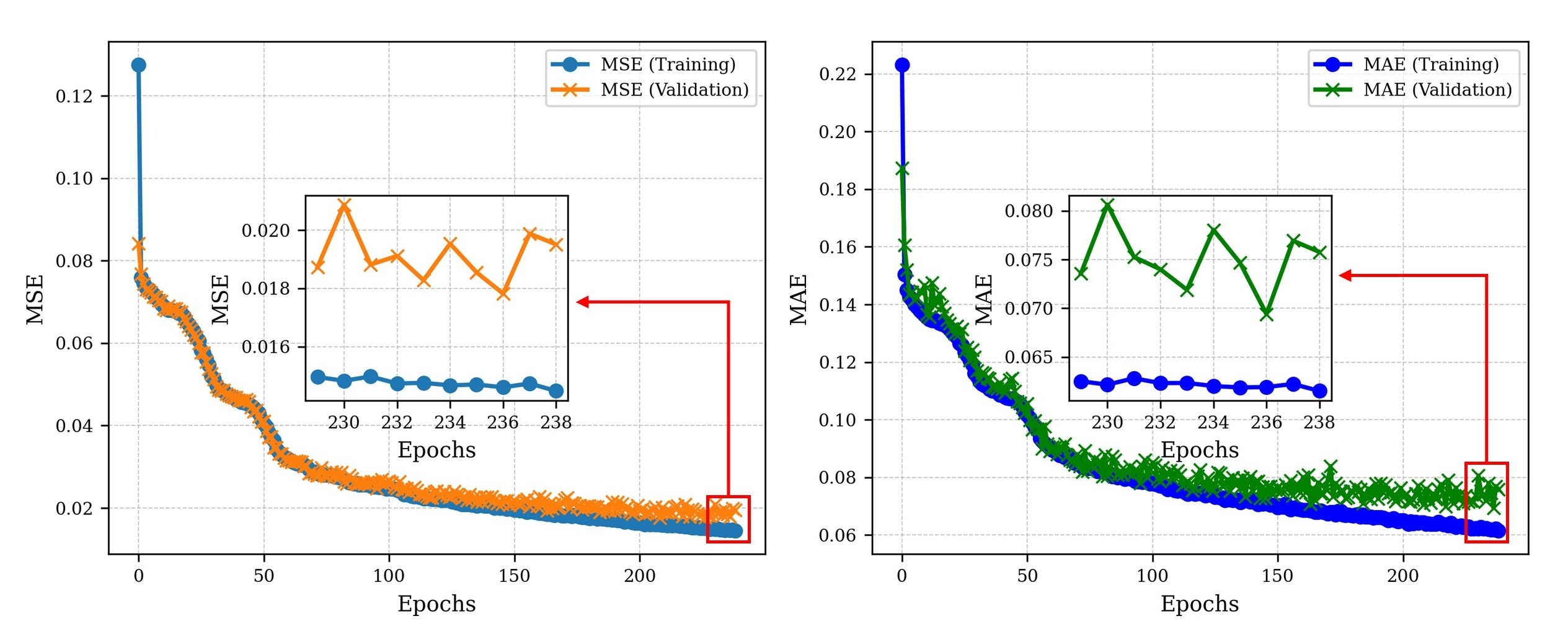}
        \caption{Learning curves of the FFM showing training and validation losses across epochs.}
        \label{figFFM_learning_curve}
    \end{subfigure}
    \vspace{0.3cm} 
    \begin{subfigure}[b]{0.8\linewidth}
        \centering
        \includegraphics[width=\linewidth]{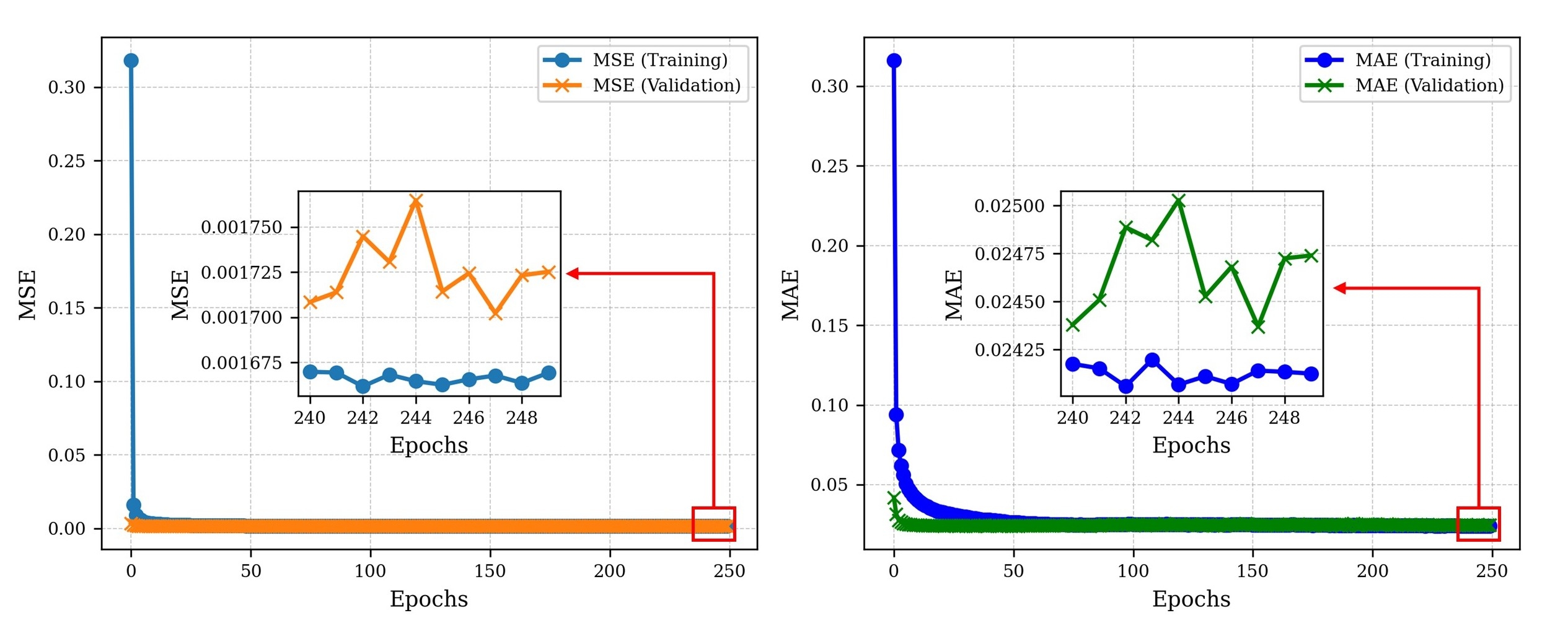}
        \caption{Learning curves of the RRM showing training and validation losses across epochs.}
        \label{fig:RRM_learning_curve}
    \end{subfigure}
    \caption{Training and validation loss curves for the HiFR\textsuperscript{2}-Net components: (a) FFM and (b) RRM.}
    \label{fig:hifr_learning}
\end{figure}

The learning behavior of the HiFR\textsuperscript{2}-Net includes three components: the FIM, the FFM, and RRM. The learning behavior of the FIM is discussed in Section~\ref{subsubsec:Learning Behavior of the FIM}. This subsection focuses on the learning behavior of the FFM and RRM, as illustrated in Figure~\ref{fig:hifr_learning}. The two-stage training process demonstrates how each sub-network contributes to refining the overall model performance.

In the first row of Figure~\ref{fig:hifr_learning}, the learning curves of the FFM are shown. Both the MSE and MAE metrics exhibit a stair-step decline throughout training. This pattern reflects stable learning behavior, characterized by phases of rapid improvement followed by periods of stagnation, possibly due to optimization plateauing. The final training and validation MSE values converge to approximately $0.0141$ and $0.0189$, respectively, while the MAE values reach around $0.0604$ (training) and $0.0746$ (validation). The small and consistent gap between training and validation metrics indicates good generalization and a reliable forward prediction baseline.

The second row of Figure~\ref{fig:hifr_learning} presents the learning curves of the RRM. The error values are very low from the beginning of training and remain nearly constant throughout the epochs. This flat curve suggests that the RRM functions primarily as a fine-tuning stage, correcting small residuals left by the FFM rather than learning from scratch. The final MSE values converge to around $0.00167$ (training) and $0.00173$ (validation), and the MAE values stabilize at approximately $0.0232$ and $0.0247$, respectively. The minimal error magnitude and narrow train-validation gap validate the RRM’s effectiveness in residual correction.

In summary, the HiFR\textsuperscript{2}-Net exhibits a two-phase learning mechanism: the FFM provides an initial coarse approximation, while the RRM incrementally refines the outputs. This hierarchical correction strategy significantly enhances accuracy and generalization compared to a single-stage network.

\subsubsection{Learning Behavior of the IRC-Net}

Based on the output of the FIM and the refinement strategy, the IRC-Net progressively improves the prediction of design parameters through a series of residual-based updates. 

\begin{figure}[t!]
    \centering
    \includegraphics[width=0.75\textwidth]{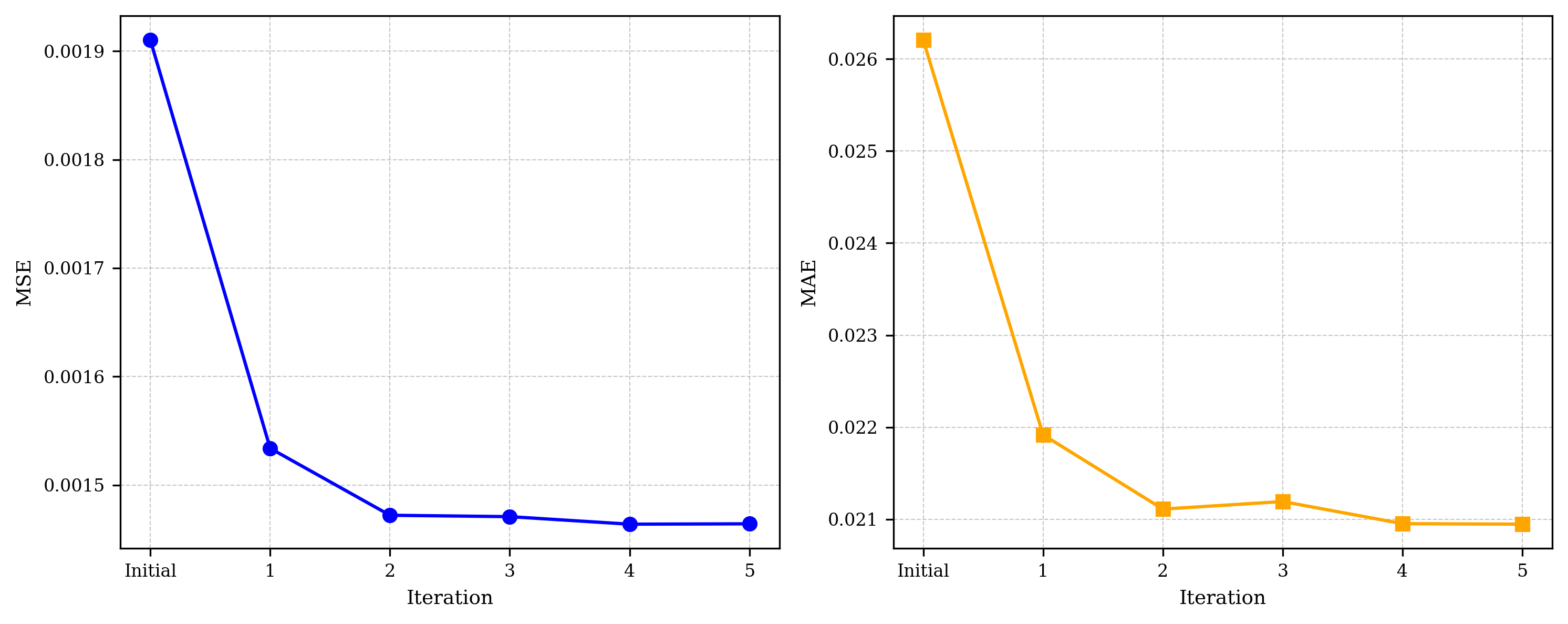}
    \caption{Reduction of prediction error over five iterations of the IRC-Net model. The initial point represents the output of the FIM. The errors are computed over the entire dataset (training, validation, and test).}
    \label{fig:irc_error_curves}
\end{figure}

Figure~\ref{fig:irc_error_curves} shows the evolution of both MSE and MAE over the course of five correction iterations. The initial point in both plots corresponds to the output of the FIM, which serves as the starting prediction \(\mathbf{P}_0\). As observed, the model achieves a sharp drop in error after the first iteration, with MSE reducing from approximately 0.00191 to 0.00154 and MAE dropping from approximately 0.0262 to 0.0219. Subsequent iterations continue to improve the performance, albeit with diminishing returns, eventually stabilizing with MSE around 0.00146 and MAE near 0.0209 by the fifth iteration.

Importantly, the reported error values are calculated throughout the data set, including training, validation, and test splits, ensuring a comprehensive assessment of the generalization and convergence behavior of the model.

This behavior illustrates that, while the FIM provides a reasonably good initialization, the iterative correction process of IRC-Net plays a critical role in enhancing prediction accuracy. The consistent convergence trend confirms the stability and effectiveness of the residual refinement framework, making it highly suitable for tackling the challenges of complex inverse design problems.

\subsection{Statistical Evaluation}

To provide a comprehensive understanding of model performance, a statistical analysis of prediction errors was conducted across the training, validation, and test subsets. Two widely used error metrics, MSE and MAE, were employed to evaluate and compare the performance of the FIM, the HiFR\textsuperscript{2}-Net, and the IRC-Net.

\begin{figure}[t!]
    \centering
    \begin{subfigure}[b]{0.6\linewidth}
        \centering
        \includegraphics[width=\linewidth]{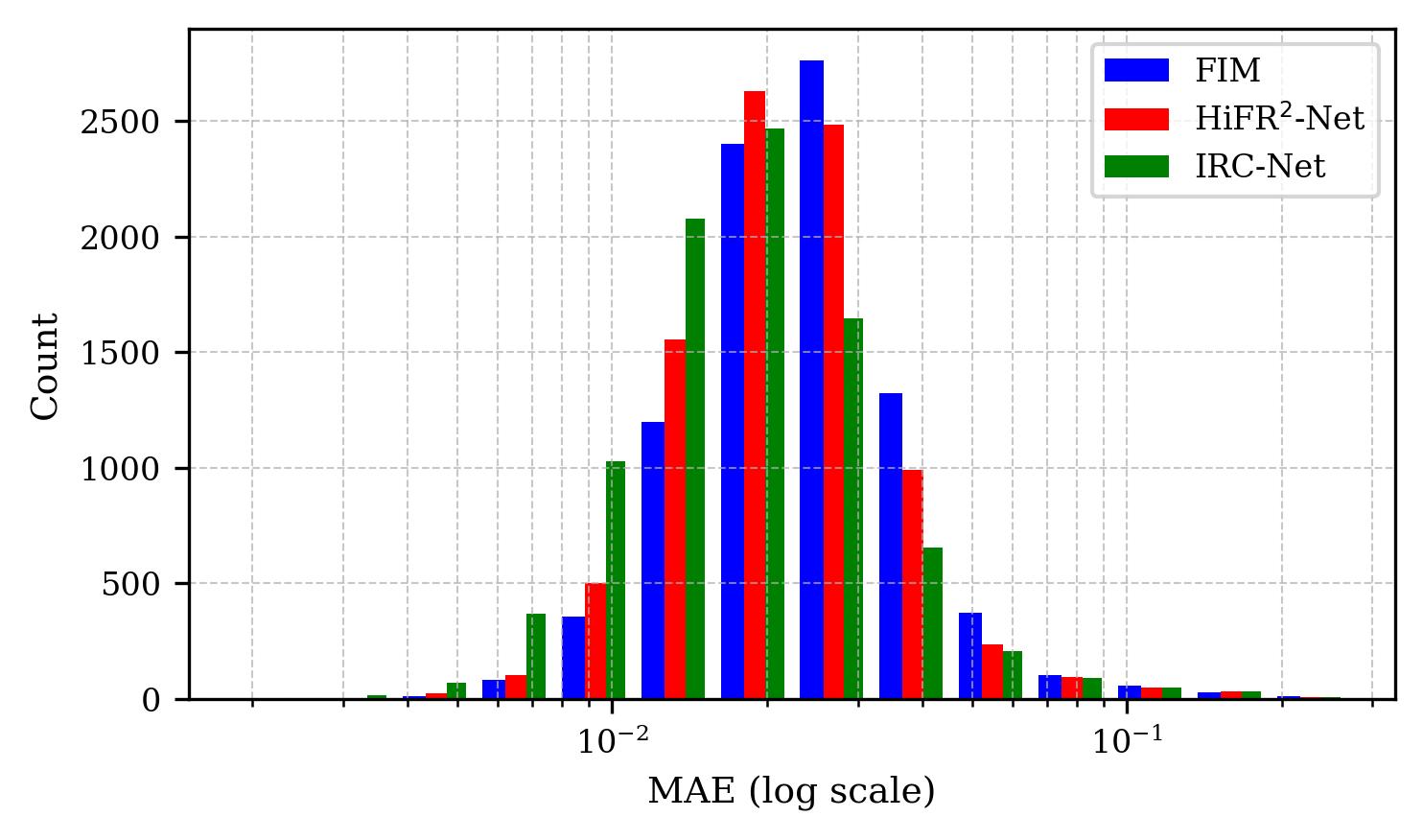}
        \caption{Histogram of MAE (log scale) for FIM, HiFR\textsuperscript{2}-Net, and IRC-Net.}
        \label{fig:mae_hist}
    \end{subfigure}

    \vspace{0.4cm}

    \begin{subfigure}[b]{0.6\linewidth}
        \centering
        \includegraphics[width=\linewidth]{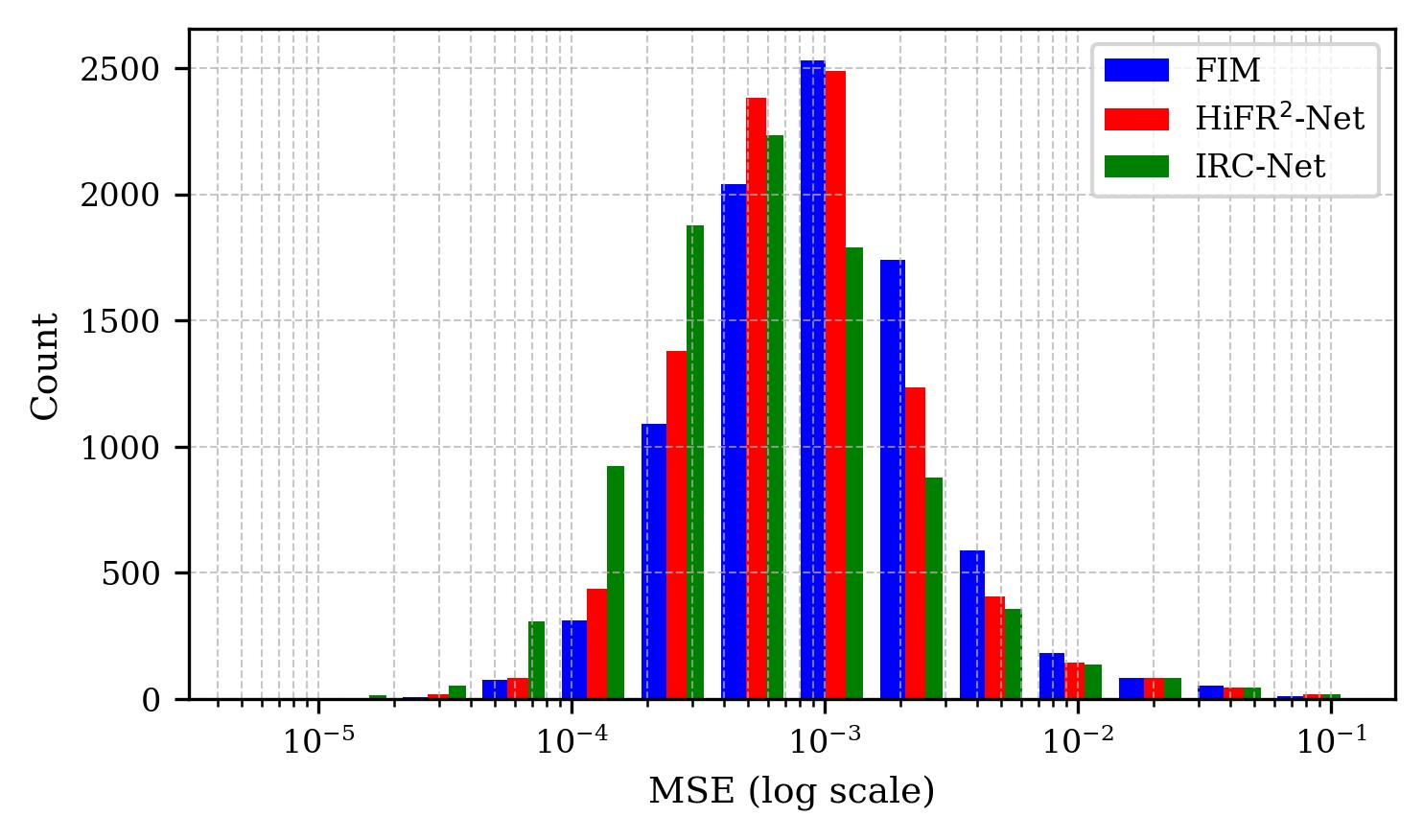}
        \caption{Histogram of MSE (log scale) for FIM, HiFR\textsuperscript{2}-Net, and IRC-Net.}
        \label{fig:mse_hist}
    \end{subfigure}

    \caption{Log-scale histograms of MAE and MSE for three network architectures.}
    \label{fig:hist_combined}
\end{figure}

Figure~\ref{fig:mae_hist} and Figure~\ref{fig:mse_hist} illustrate the distribution of MAE and MSE values, respectively, on a logarithmic scale. These histograms provide insights not only into the average performance of each model but also into their consistency and robustness.

As shown in Figure~\ref{fig:mae_hist}, the MAE histogram reveals that the IRC-Net (green) achieves the lowest and most tightly clustered errors among all models. Its distribution peaks at a lower MAE value and exhibits minimal spread, indicating a high degree of consistency and fine-grained accuracy in its predictions. The HiFR\textsuperscript{2}-Net (red) also demonstrates improved performance over the baseline FIM (blue), with a shifted peak toward lower errors and a narrower spread. In contrast, the FIM shows a broader distribution with a heavier tail on the right, reflecting its relatively higher prediction variability and larger occurrence of high-error instances.

Similarly, the MSE distribution in Figure~\ref{fig:mse_hist} supports these observations. The IRC-Net maintains a compact error distribution with the majority of its predictions concentrated around the lowest MSE values. The HiFR\textsuperscript{2}-Net again performs better than the FIM by achieving a more left-shifted peak and reduced variance. Notably, the logarithmic scale highlights the IRC-Net’s effectiveness in minimizing even small residual errors, a critical requirement in high-precision inverse design applications.

Collectively, the statistical evaluations and comparative visualizations confirm the superiority of the proposed IRC-Net architecture. While the FIM provides a reasonable initial estimate, it is the combination of coarse-to-fine corrections in the HiFR\textsuperscript{2}-Net and the iterative residual refinement in the IRC-Net that results in highly accurate and reliable inverse design predictions. The IRC-Net’s ability to correct residual errors through multiple refinement stages proves particularly effective, enabling it to deliver more precise and robust performance across the entire dataset.

\subsection{Parametric Evaluation and Experimental Validation} 

To comprehensively evaluate the performance of the proposed \emph{IRC-Net} framework, its capability to accurately predict the design parameters of multimode SIW resonators is first assessed. The model’s predictions are compared against the ground truth for the structure illustrated in Figure~\ref{fig:sparams}, which was randomly selected from the test dataset. The predicted design parameters from the three models, FIM, HiFR\textsuperscript{2}-Net, and IRC-Net are compared with the true target values in \hyperref[tab:predictions]{Table 6}, where the IRC-Net demonstrates the closest match to the target parameters, highlighting its superior accuracy in inverse modeling.

\begin{table}[htbp]
\begin{minipage}[t]{\textwidth}
\raggedright
{\small \textbf{Table 6} \label{tab:predictions} \\ 
Comparison between model predictions and ground truth for various parameters.}
\vspace{0.8em}
\end{minipage}

\centering
{\small
\begin{tabular}{ccccc}
\toprule
\textbf{Parameters} & \textbf{FIM-Net} & \textbf{HiFR\textsuperscript{2}-Net} & \textbf{IRC-Net} & \textbf{Ground Truth} \\
\midrule
D1 & 5.5871 & 5.8126 & 5.5745 & 5.5000 \\
D2 & 7.8192 & 7.9293 & 7.9494 & 8.0000 \\
R1 & 0.2118 & 0.2136 & 0.2174 & 0.2000 \\
R2 & 0.3846 & 0.3723 & 0.4084 & 0.4000 \\
R3 & 0.8210 & 0.8065 & 0.8047 & 0.8000 \\
G  & 25.7656 & 26.1080 & 26.0227 & 26.0000 \\
\bottomrule
\end{tabular}
}
\end{table}

\begin{figure}[t!]
    \centering
    \includegraphics[width=0.85\linewidth]{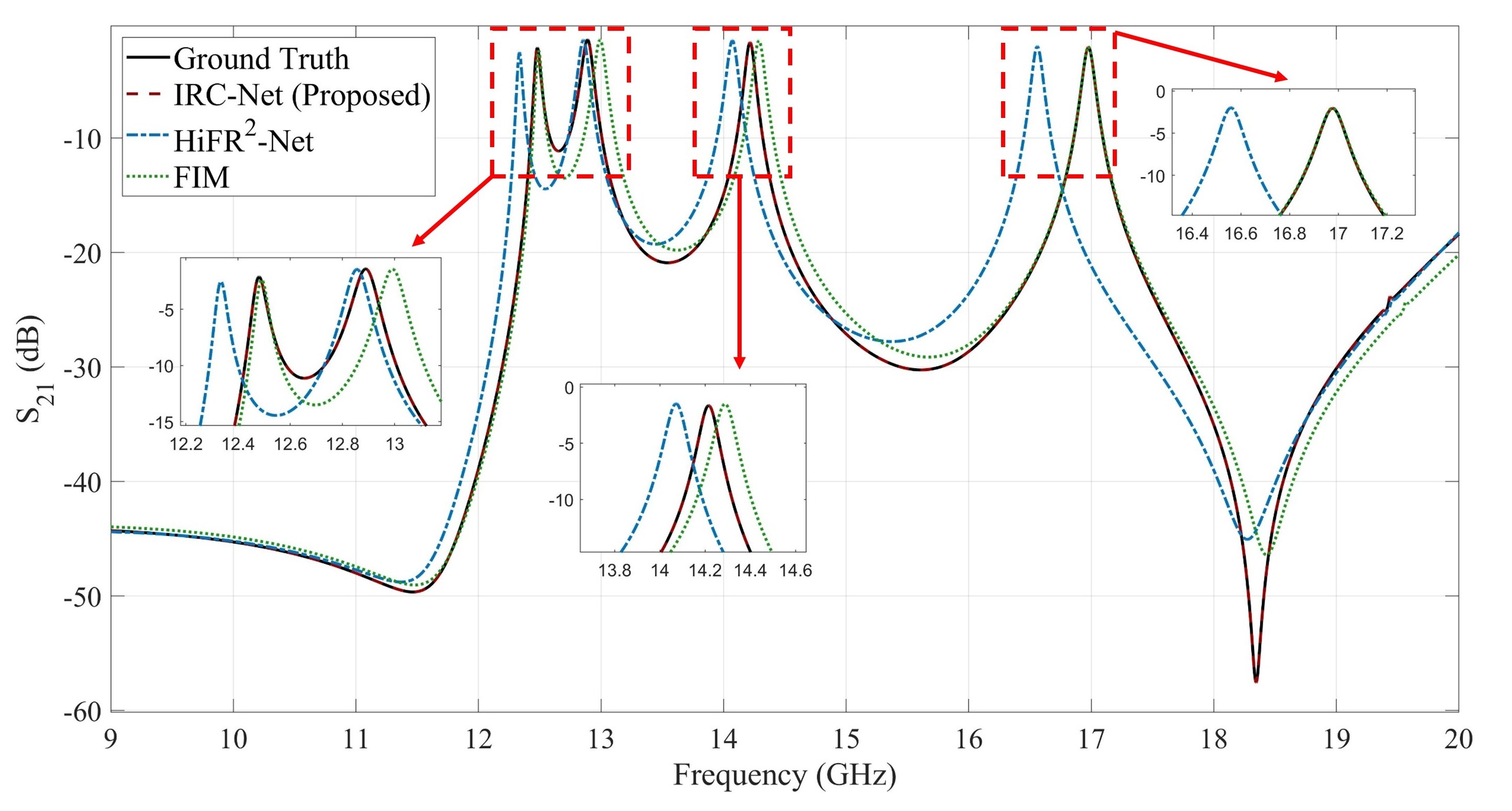}
    \caption{Simulated \( S_{21} \) parameter responses using predicted design parameters from different models compared with the ground truth.}
    \label{fig:simulated_sparams}
\end{figure}

\hyperref[tab:predictions]{Table 6} clearly demonstrates that the \emph{IRC-Net} model provides significantly more accurate predictions of the design parameters compared to the \emph{FIM} and \emph{HiFR\textsuperscript{2}-Net} models. To further validate these findings, simulations were performed using the predicted design parameters.  Only the resulting \( S_{21} \) responses are presented for streamlined comparison, as shown in Figure~\ref{fig:simulated_sparams}. The simulations provide an empirical validation by comparing electromagnetic performance against the ground truth, and the \emph{IRC-Net} yields an $S_{21}$ response closely matching the ground truth across the entire frequency range, outperforming both the \emph{FIM} and \emph{HiFR\textsuperscript{2}-Net} models.

\begin{figure}[t!]
	\centering
	
	\begin{subfigure}[b]{0.3\linewidth}
		\centering
		\includegraphics[width=\linewidth]{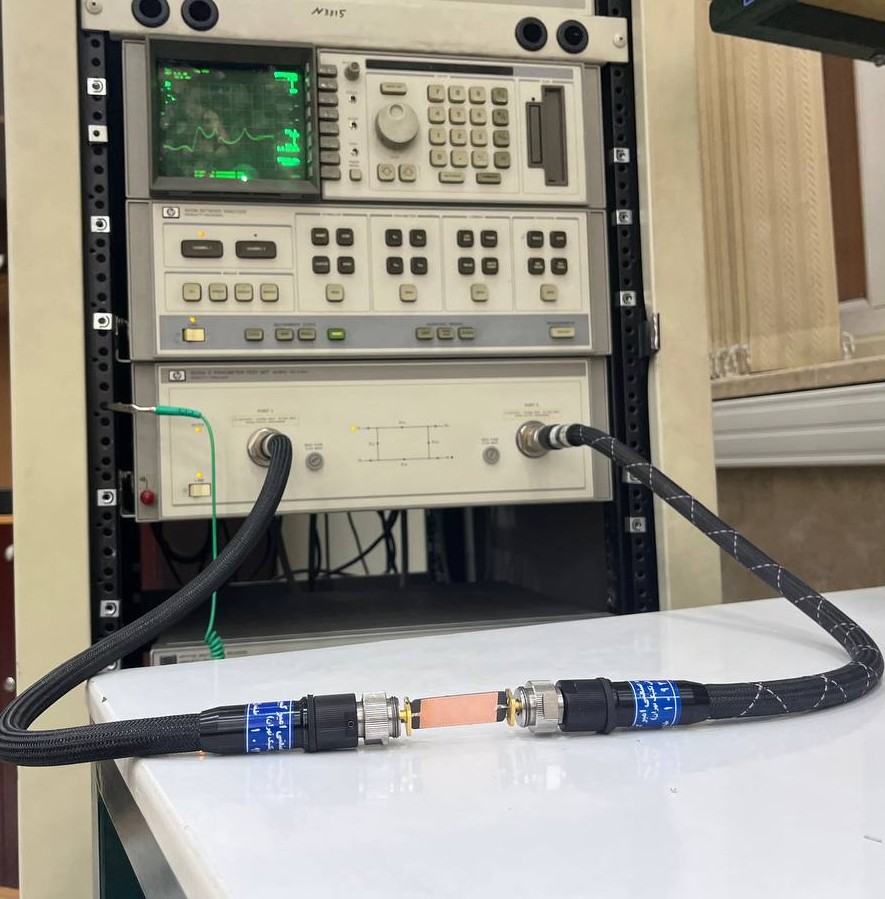}
		\caption{Fabricated SIW structure exhibiting three in-band resonances.}
		\label{fig:meas_setupa}
	\end{subfigure}
	\hspace{0.02\linewidth}
	\begin{subfigure}[b]{0.3\linewidth}
		\centering
		\includegraphics[width=\linewidth]{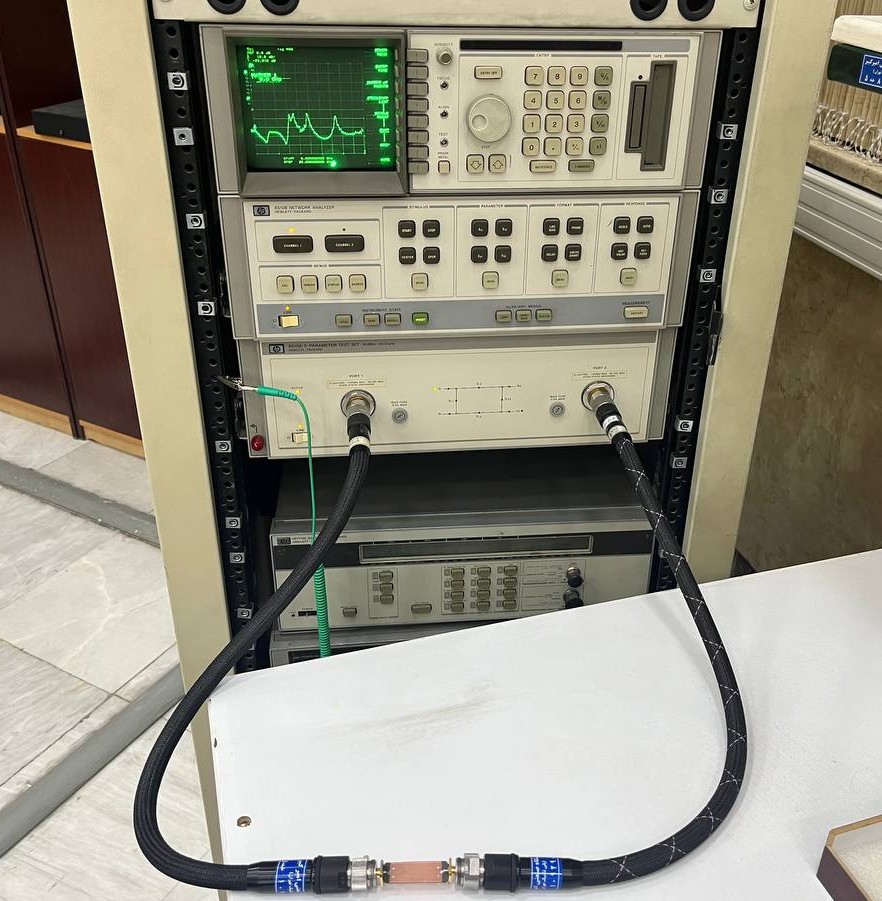}
		\caption{Fabricated SIW structure exhibiting four in-band resonances.}
		\label{fig:meas_setupb}
	\end{subfigure}
	
	\caption{Measurement setup for the fabricated SIW structures using an HP8510 VNA.}
	\label{fig:meas_setup0}
\end{figure}

To evaluate the practical effectiveness of the proposed IRC-Net framework, multi-mode SIW resonator structures were selected for fabrication and experimental validation. The first design, which exhibits a four-resonance response, was developed using conventional full-wave electromagnetic simulation techniques. In contrast, the second structure, which features three in-band resonances, was designed entirely by the trained IRC-Net model based on the desired frequency response. Both structures were fabricated using standard PCB manufacturing processes on the RT/Duroid 5880 substrate with a thickness of 20 mils. The S-parameters of the fabricated devices were measured using an HP8510 vector network analyzer, as illustrated in Figure~\ref{fig:meas_setup0}.

The input signal was generated by an external frequency synthesizer, and the input power level was set to $-30\,\mathrm{dBm}$ to ensure that the input mixers operate within their linear range. The frequency sweep was performed from 9\,GHz to 20\,GHz.

\begin{figure}[ht!]
  \centering
  \begin{subfigure}[b]{0.85\linewidth}
    \centering
    \includegraphics[width=\linewidth,keepaspectratio]{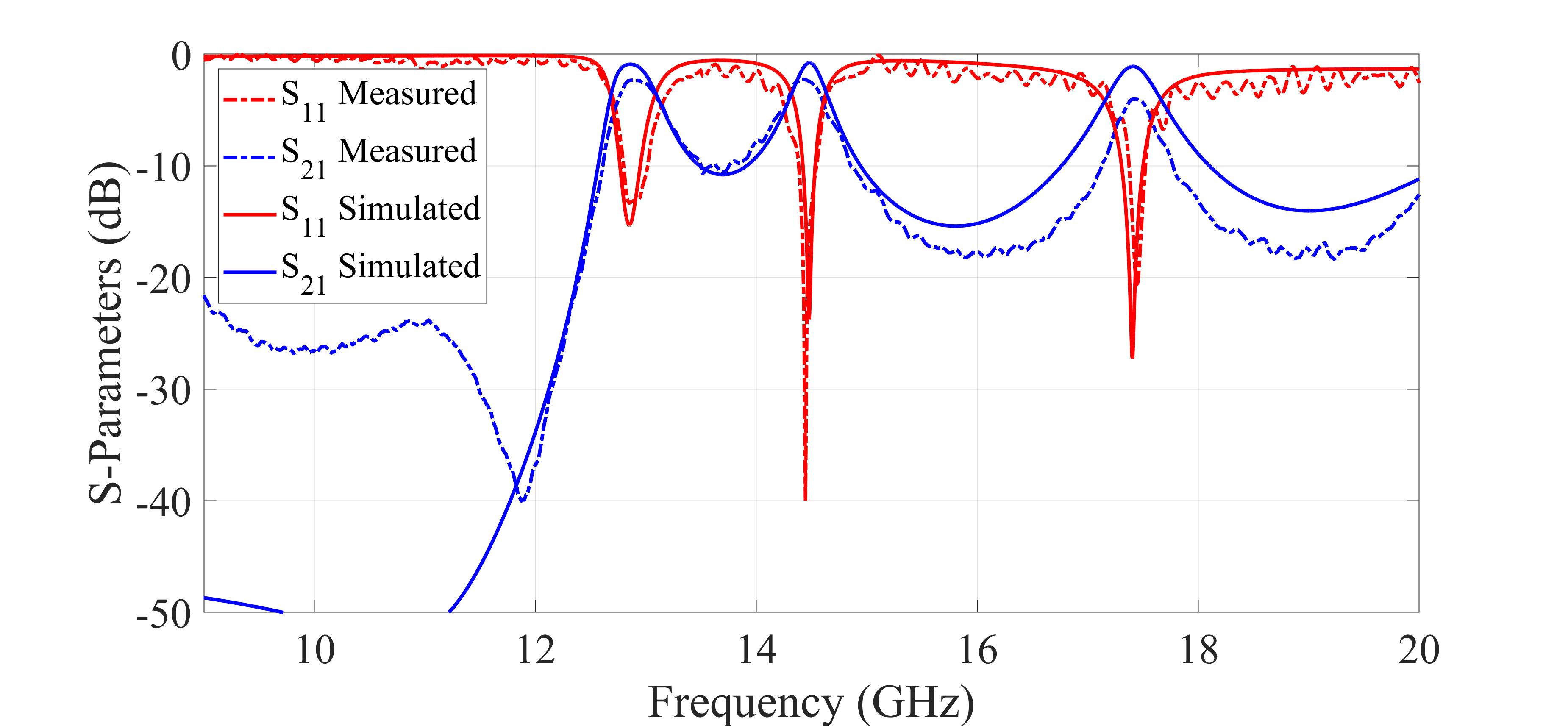}
    \caption{Measured vs. predicted S-parameters for the structure with three in-band resonances.}
    \label{fig:3res}
  \end{subfigure}

  \vspace{0.4cm}

  \begin{subfigure}[b]{0.85\linewidth}
    \centering
    \includegraphics[width=\linewidth,keepaspectratio]{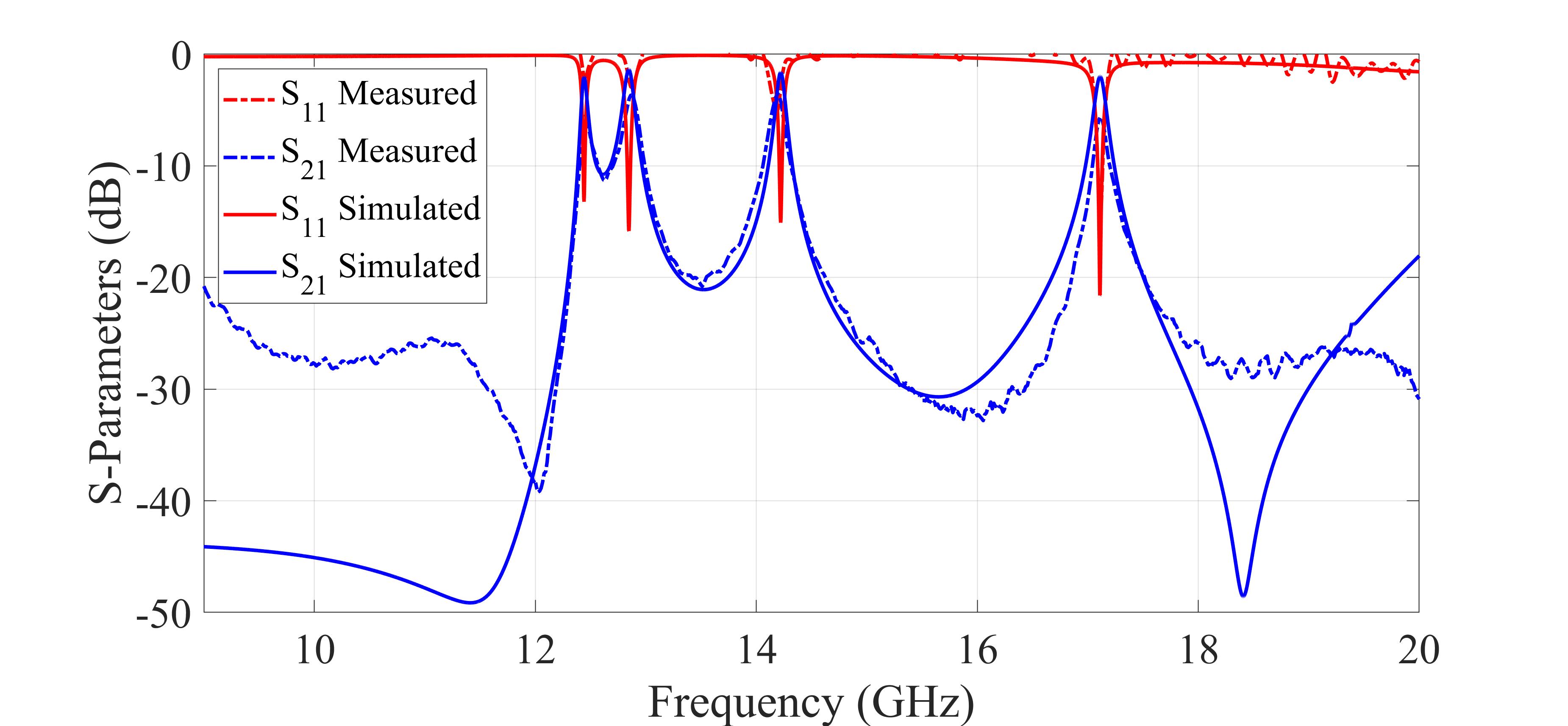}
    \caption{Measured vs. simulated S-parameters for the structure with four in-band resonances.}
    \label{fig:4res}
  \end{subfigure}

  \caption{Comparison between the measured, simulated and predicted S-parameters for the fabricated SIW structures: (a) three-resonance structure, (b) four-resonance structure.}
  \label{fig:comparison}
\end{figure}

Figure~\ref{fig:comparison} presents a comparison between the measured, simulated, and predicted S-parameters. As observed, there is an excellent agreement between the fabricated and predicted results, confirming the effectiveness of the proposed model. This strong correlation indicates that the \emph{IRC-Net} framework can be reliably extended to more complex passive microwave resonant structures involving higher design freedom and a larger number of geometric parameters. In summary, the fabrication and measurement results of both the three- and four-resonance SIW prototypes validate the high accuracy and practical applicability of the proposed framework, demonstrating its capability to guide the design of complex multi-mode resonators with confidence.

\section{Conclusion}
\label{sec:conclusion}
We have presented a three-stage progressive deep learning framework for the inverse design of multi‑mode SIW filters, comprising a Feedforward Inverse Model, a ~Hybrid ~Inverse‑Forward ~Residual ~Refinement ~Network, and an Iterative Residual Correction Network (IRC‑Net). Experimental validation on fabricated prototypes—including geometries outside the training set—demonstrated that IRC‑Net consistently outperforms the baseline models, reducing mean squared error from 0.00191 to 0.00146 and mean absolute error from 0.0262 to 0.0209, while enabling sub‑millisecond prediction times. This approach significantly reduces dependence on computationally intensive electromagnetic simulations and facilitates rapid prototyping of complex microwave components.

Although the demonstrated results focus on symmetric filters with up to four resonators, the modular and iterative architecture of IRC‑Net holds promise for extension to higher-order and asymmetric SIW configurations, as well as broader applications across microwave and millimeter-wave domains.
However, two main limitations remain. First, the generalization capability of the model across more diverse structures—such as those with varying topology, material properties, or irregular geometries—has not yet been fully explored. Future work should investigate training on larger and more heterogeneous datasets to improve robustness and applicability to a wider range of practical designs.

Second, like most data-driven models, the proposed framework operates as a black box, offering limited insight into the underlying physical reasoning behind its predictions. To address this, future research may incorporate concepts from explainable artificial intelligence (XAI) to improve interpretability and trustworthiness, potentially revealing new physics-informed design principles embedded within the network.

\section*{CRediT authorship contribution statement}

\textbf{Mohammad Mashayekhi:} Conceptualization, Methodology, Software, Data curation, Project administration, Writing – original draft. \textbf{Kamran Salehian:} Conceptualization, SIW structure design, Fabrication of SIW, Figure preparation for SIW design, Writing – review \& editing. \textbf{Abbas Ozgoli:} Fabrication and measurement of SIW. \textbf{Saeed Abdollahi:} Writing – introduction. \textbf{Abdolali Abdipour:} Supervision. \textbf{Ahmed A. Kishk:} Supervision, Project administration, Writing – review \& editing.

\section*{Funding}
This work is self-funded research. NO Government, External organizations are involved with financial assistance.

\section*{Declaration of competing interest}

The authors declare that they have no known competing financial interests or personal relationships that could have appeared to influence the work reported in this paper.

\section*{Data availability}

The dataset generated and analyzed during the current study is not publicly available due to its integration into ongoing research projects and future planned publications.

\bibliographystyle{elsarticle-num}
\bibliography{references}

\begin{thebibliography}{10}
\expandafter\ifx\csname url\endcsname\relax
  \def\url#1{\texttt{#1}}\fi
\expandafter\ifx\csname urlprefix\endcsname\relax\def\urlprefix{URL }\fi
\expandafter\ifx\csname href\endcsname\relax
  \def\href#1#2{#2} \def\path#1{#1}\fi

\bibitem{10486926}
M.~Qin, Z.~Li, Z.~Ren, P.~Liu, L.~Liao, X.~Qiu, Z.~Li, Varactor-based
  continuously tunable microstrip bandpass filters: A review, issues and future
  trends, IEEE Access 12 (2024) 57443--57457.
\newblock \href {https://doi.org/10.1109/ACCESS.2024.3383788}
  {\path{doi:10.1109/ACCESS.2024.3383788}}.

\bibitem{10417067}
P.~Vryonides, S.~Arain, A.~Quddious, D.~Psychogiou, S.~Nikolaou, A new class of
  high-selectivity bandpass filters with constant bandwidth and 5:1 bandwidth
  tuning ratio, IEEE Access 12 (2024) 16489--16497.
\newblock \href {https://doi.org/10.1109/ACCESS.2024.3358677}
  {\path{doi:10.1109/ACCESS.2024.3358677}}.

\bibitem{ZAKHAROV2024155131}
A.~V. Zakharov, S.~M. Litvintsev,
  \href{https://www.sciencedirect.com/science/article/pii/S1434841124000165}{Transmission
  line bandpass filters with multiple attenuation poles and small number of
  resonators}, AEU - International Journal of Electronics and Communications
  176 (2024) 155131.
\newblock \href {https://doi.org/https://doi.org/10.1016/j.aeue.2024.155131}
  {\path{doi:https://doi.org/10.1016/j.aeue.2024.155131}}.
\newline\urlprefix\url{https://www.sciencedirect.com/science/article/pii/S1434841124000165}

\bibitem{7054566}
R.~Gómez-García, A.~C. Guyette, Reconfigurable multi-band microwave filters,
  IEEE Transactions on Microwave Theory and Techniques 63~(4) (2015)
  1294--1307.
\newblock \href {https://doi.org/10.1109/TMTT.2015.2405066}
  {\path{doi:10.1109/TMTT.2015.2405066}}.

\bibitem{s22051961}
V.~V. Stanovov, S.~A. Khodenkov, A.~M. Popov, L.~A. Kazakovtsev,
  \href{https://www.mdpi.com/1424-8220/22/5/1961}{The automatic design of
  multimode resonator topology with evolutionary algorithms}, Sensors 22~(5)
  (2022).
\newblock \href {https://doi.org/10.3390/s22051961}
  {\path{doi:10.3390/s22051961}}.
\newline\urlprefix\url{https://www.mdpi.com/1424-8220/22/5/1961}

\bibitem{electronics10222853}
S.~Palanisamy, B.~Thangaraju, O.~I. Khalaf, Y.~Alotaibi, S.~Alghamdi,
  \href{https://www.mdpi.com/2079-9292/10/22/2853}{Design and synthesis of
  multi-mode bandpass filter for wireless applications}, Electronics 10~(22)
  (2021).
\newblock \href {https://doi.org/10.3390/electronics10222853}
  {\path{doi:10.3390/electronics10222853}}.
\newline\urlprefix\url{https://www.mdpi.com/2079-9292/10/22/2853}

\bibitem{7859469}
J.~Xu, W.~Hong, H.~Zhang, H.~Tang, Compact bandpass filter with multiple
  coupling paths in limited space for ku-band application, IEEE Microwave and
  Wireless Components Letters 27~(3) (2017) 251--253.
\newblock \href {https://doi.org/10.1109/LMWC.2017.2661970}
  {\path{doi:10.1109/LMWC.2017.2661970}}.

\bibitem{10107477}
T.~E. Humphreys, P.~A. Iannucci, Z.~M. Komodromos, A.~M. Graff, Signal
  structure of the starlink ku-band downlink, IEEE Transactions on Aerospace
  and Electronic Systems 59~(5) (2023) 6016--6030.
\newblock \href {https://doi.org/10.1109/TAES.2023.3268610}
  {\path{doi:10.1109/TAES.2023.3268610}}.

\bibitem{10896684}
S.~Abdollahi, A.~Bakhtafrouz, M.~Maddahali, Design of a beamforming tri-band
  rectifier for rf energy harvesting, IEEE Access 13 (2025) 34802--34811.
\newblock \href {https://doi.org/10.1109/ACCESS.2025.3544146}
  {\path{doi:10.1109/ACCESS.2025.3544146}}.

\bibitem{10038507}
Z.~Wu, M.~Wang, X.~Liao, Y.~Pu, L.~Wang, W.~Jiang, J.~Wang, Y.~Luo, Oversized
  multimode waveguide filter for circular te$_{01}$ mode high power
  transmission lines, IEEE Transactions on Microwave Theory and Techniques
  71~(8) (2023) 3552--3560.
\newblock \href {https://doi.org/10.1109/TMTT.2023.3240513}
  {\path{doi:10.1109/TMTT.2023.3240513}}.

\bibitem{FARASHAHI201815}
M.~Farashahi, E.~Zareian-Jahromi, R.~Basiri,
  \href{https://www.sciencedirect.com/science/article/pii/S143484111830195X}{A
  compact semi-open wideband siw horn antenna for k/ku band applications}, AEU
  - International Journal of Electronics and Communications 92 (2018) 15--20.
\newblock \href {https://doi.org/https://doi.org/10.1016/j.aeue.2018.05.012}
  {\path{doi:https://doi.org/10.1016/j.aeue.2018.05.012}}.
\newline\urlprefix\url{https://www.sciencedirect.com/science/article/pii/S143484111830195X}

\bibitem{10648671}
A.~Alam, M.~S. Alam, K.~Almuhanna, A.~Shamim, A critical review of interconnect
  options for siw technologies, IEEE Access 12 (2024) 122902--122917.
\newblock \href {https://doi.org/10.1109/ACCESS.2024.3450111}
  {\path{doi:10.1109/ACCESS.2024.3450111}}.

\bibitem{Darling2025}
J.~S. Darling, V.~Guruviah, R.~P. Dwivedi,
  \href{https://doi.org/10.1002/dac.6036}{Siw technology for 5g antenna
  applications and beyond—a critical review}, International Journal of
  Communication Systems 38~(2) (2025) e6036.
\newblock \href {https://doi.org/10.1002/dac.6036}
  {\path{doi:10.1002/dac.6036}}.
\newline\urlprefix\url{https://doi.org/10.1002/dac.6036}

\bibitem{9760077}
T.~Iye, P.~van Wyk, T.~Matsumoto, Y.~Susukida, S.~Takaya, Y.~Fujii, Neural
  network-based phase estimation for antenna array using radiation power
  pattern, IEEE Antennas and Wireless Propagation Letters 21~(7) (2022)
  1348--1352.
\newblock \href {https://doi.org/10.1109/LAWP.2022.3167697}
  {\path{doi:10.1109/LAWP.2022.3167697}}.

\bibitem{Mashayekhi2024}
M.~Mashayekhi, H.~Soleimani,
  \href{https://ietresearch.onlinelibrary.wiley.com/doi/abs/10.1049/mia2.70018}{Fundamental
  and harmonic beamforming of desire time-modulated planar arrays with deep
  learning}, IET Microwaves, Antennas \& Propagation 19~(1) (2025).
\newblock \href
  {http://arxiv.org/abs/https://ietresearch.onlinelibrary.wiley.com/doi/pdf/10.1049/mia2.70018}
  {\path{arXiv:https://ietresearch.onlinelibrary.wiley.com/doi/pdf/10.1049/mia2.70018}},
  \href {https://doi.org/https://doi.org/10.1049/mia2.70018}
  {\path{doi:https://doi.org/10.1049/mia2.70018}}.
\newline\urlprefix\url{https://ietresearch.onlinelibrary.wiley.com/doi/abs/10.1049/mia2.70018}

\bibitem{Ghorbani2021}
F.~Ghorbani, J.~Shabanpour, S.~Beyraghi, H.~Soleimani, H.~Oraizi, M.~Soleimani,
  \href{https://doi.org/10.1007/s00339-021-05030-6}{A deep learning approach
  for inverse design of the metasurface for dual-polarized waves}, Applied
  Physics A 127~(11) (2021) 869.
\newblock \href {https://doi.org/10.1007/s00339-021-05030-6}
  {\path{doi:10.1007/s00339-021-05030-6}}.
\newline\urlprefix\url{https://doi.org/10.1007/s00339-021-05030-6}

\bibitem{Ghorbani2021b}
F.~Ghorbani, S.~Beyraghi, J.~Shabanpour, H.~Oraizi, H.~Soleimani, M.~Soleimani,
  \href{https://doi.org/10.1038/s41598-021-86588-2}{Deep neural network-based
  automatic metasurface design with a wide frequency range}, Scientific Reports
  11~(1) (2021) 7102.
\newblock \href {https://doi.org/10.1038/s41598-021-86588-2}
  {\path{doi:10.1038/s41598-021-86588-2}}.
\newline\urlprefix\url{https://doi.org/10.1038/s41598-021-86588-2}

\bibitem{Mashayekhi2023}
M.~Mashayekhi, P.~Kabiri, A.~S. Nooramin, M.~Soleimani,
  \href{https://doi.org/10.1038/s41598-023-35036-4}{A reconfigurable graphene
  patch antenna inverse design at terahertz frequencies}, Scientific Reports
  13~(1) (2023) 8369.
\newblock \href {https://doi.org/10.1038/s41598-023-35036-4}
  {\path{doi:10.1038/s41598-023-35036-4}}.
\newline\urlprefix\url{https://doi.org/10.1038/s41598-023-35036-4}

\bibitem{10130105}
A.~Gupta, E.~A. Karahan, C.~Bhat, K.~Sengupta, U.~K. Khankhoje, Tandem neural
  network based design of multiband antennas, IEEE Transactions on Antennas and
  Propagation 71~(8) (2023) 6308--6317.
\newblock \href {https://doi.org/10.1109/TAP.2023.3276524}
  {\path{doi:10.1109/TAP.2023.3276524}}.

\bibitem{10702499}
L.~Ahmadi, A.~A. Shishegar, A physics-based deep learning to extend born
  approximation validity to strong scatterers, IEEE Transactions on Antennas
  and Propagation 72~(12) (2024) 9392--9400.
\newblock \href {https://doi.org/10.1109/TAP.2024.3467700}
  {\path{doi:10.1109/TAP.2024.3467700}}.

\bibitem{Beyraghi2023}
S.~Beyraghi, F.~Ghorbani, J.~Shabanpour, M.~E. Lajevardi, V.~Nayyeri, P.-Y.
  Chen, O.~M. Ramahi,
  \href{https://doi.org/10.1038/s41598-023-44131-5}{Microwave bone fracture
  diagnosis using deep neural network}, Scientific Reports 13~(1) (2023) 16957.
\newblock \href {https://doi.org/10.1038/s41598-023-44131-5}
  {\path{doi:10.1038/s41598-023-44131-5}}.
\newline\urlprefix\url{https://doi.org/10.1038/s41598-023-44131-5}

\bibitem{10472507}
M.~R. Khan, C.~L. Zekios, S.~Bhardwaj, S.~V. Georgakopoulos, A deep learning
  convolutional neural network for antenna near-field prediction and surrogate
  modeling, IEEE Access 12 (2024) 39737--39747.
\newblock \href {https://doi.org/10.1109/ACCESS.2024.3377219}
  {\path{doi:10.1109/ACCESS.2024.3377219}}.

\bibitem{USTUN201954}
D.~Ustun, A.~Toktas, A.~Akdagli,
  \href{https://www.sciencedirect.com/science/article/pii/S1434841118332205}{Deep
  neural network–based soft computing the resonant frequency of e–shaped
  patch antennas}, AEU - International Journal of Electronics and
  Communications 102 (2019) 54--61.
\newblock \href {https://doi.org/https://doi.org/10.1016/j.aeue.2019.02.011}
  {\path{doi:https://doi.org/10.1016/j.aeue.2019.02.011}}.
\newline\urlprefix\url{https://www.sciencedirect.com/science/article/pii/S1434841118332205}

\bibitem{10944803}
G.~Chaudhary, Y.~Meng, D.~Sun~Park, Y.~Jeong, Attention-based deep neural
  network for high-dimensional microwave modeling of non-reciprocal bandpass
  filters, IEEE Access 13 (2025) 56220--56236.
\newblock \href {https://doi.org/10.1109/ACCESS.2025.3555583}
  {\path{doi:10.1109/ACCESS.2025.3555583}}.

\bibitem{4470587}
H.~Kabir, Y.~Wang, M.~Yu, Q.-J. Zhang, Neural network inverse modeling and
  applications to microwave filter design, IEEE Transactions on Microwave
  Theory and Techniques 56~(4) (2008) 867--879.
\newblock \href {https://doi.org/10.1109/TMTT.2008.919078}
  {\path{doi:10.1109/TMTT.2008.919078}}.

\bibitem{8798884}
J.~Jin, C.~Zhang, F.~Feng, W.~Na, J.~Ma, Q.-J. Zhang, Deep neural network
  technique for high-dimensional microwave modeling and applications to
  parameter extraction of microwave filters, IEEE Transactions on Microwave
  Theory and Techniques 67~(10) (2019) 4140--4155.
\newblock \href {https://doi.org/10.1109/TMTT.2019.2932738}
  {\path{doi:10.1109/TMTT.2019.2932738}}.

\bibitem{9076290}
G.~Pan, Y.~Wu, M.~Yu, L.~Fu, H.~Li, Inverse modeling for filters using a
  regularized deep neural network approach, IEEE Microwave and Wireless
  Components Letters 30~(5) (2020) 457--460.
\newblock \href {https://doi.org/10.1109/LMWC.2020.2986156}
  {\path{doi:10.1109/LMWC.2020.2986156}}.

\bibitem{Karahan2024}
E.~A. Karahan, Z.~Liu, A.~Gupta, Z.~Shao, J.~Zhou, U.~Khankhoje, K.~Sengupta,
  \href{https://doi.org/10.1038/s41467-024-54178-1}{Deep-learning enabled
  generalized inverse design of multi-port radio-frequency and sub-terahertz
  passives and integrated circuits}, Nature Communications 15~(1) (2024) 10734.
\newblock \href {https://doi.org/10.1038/s41467-024-54178-1}
  {\path{doi:10.1038/s41467-024-54178-1}}.
\newline\urlprefix\url{https://doi.org/10.1038/s41467-024-54178-1}

\bibitem{10867510}
S.~Luo, J.~Ma, S.~Dang, A.~Austin, Deep learning-enabled rapid optimization for
  microwave filter design, in: 2024 IEEE Asia-Pacific Microwave Conference
  (APMC), 2024, pp. 1287--1289.
\newblock \href {https://doi.org/10.1109/APMC60911.2024.10867510}
  {\path{doi:10.1109/APMC60911.2024.10867510}}.

\bibitem{9184562}
G.-y. Du, L.~Jin, Design of siw filter based on the equivalent de-embedding
  technique and inverse neural network, in: 2020 IEEE 63rd International
  Midwest Symposium on Circuits and Systems (MWSCAS), 2020, pp. 407--410.
\newblock \href {https://doi.org/10.1109/MWSCAS48704.2020.9184562}
  {\path{doi:10.1109/MWSCAS48704.2020.9184562}}.

\bibitem{9223952}
H.~Yu, H.~M. Torun, M.~U. Rehman, M.~Swaminathan, Design of siw filters in
  d-band using invertible neural nets, in: 2020 IEEE/MTT-S International
  Microwave Symposium (IMS), 2020, pp. 72--75.
\newblock \href {https://doi.org/10.1109/IMS30576.2020.9223952}
  {\path{doi:10.1109/IMS30576.2020.9223952}}.

\bibitem{Soundarya2023}
G.~Soundarya, N.~Gunavathi,
  \href{https://doi.org/10.1080/03772063.2023.2177202}{Ku-band siw filter with
  high fractional bandwidth optimized using feed forward back propagation ann},
  IETE Journal of Research 70~(2) (2023) 1272--1282.
\newblock \href {https://doi.org/10.1080/03772063.2023.2177202}
  {\path{doi:10.1080/03772063.2023.2177202}}.
\newline\urlprefix\url{https://doi.org/10.1080/03772063.2023.2177202}

\bibitem{10354047}
J.~Zhang, J.~Shao, B.-Z. Wang, R.~Wang, Inverse design of siw devices based on
  the internal multiport method, IEEE Microwave and Wireless Technology Letters
  34~(2) (2024) 171--174.
\newblock \href {https://doi.org/10.1109/LMWT.2023.3339697}
  {\path{doi:10.1109/LMWT.2023.3339697}}.

\bibitem{Stetter1978}
H.~J. Stetter, \href{https://doi.org/10.1007/BF01432879}{The defect correction
  principle and discretization methods}, Numerische Mathematik 29~(4) (1978)
  425--443.
\newblock \href {https://doi.org/10.1007/BF01432879}
  {\path{doi:10.1007/BF01432879}}.
\newline\urlprefix\url{https://doi.org/10.1007/BF01432879}

\bibitem{KAUCHER1984142}
E.~W. Kaucher, W.~L. Miranker,
  \href{https://www.sciencedirect.com/science/article/pii/B9780124020207500116}{Chapter
  5 - iterative residual correction}, in: E.~W. Kaucher, W.~L. Miranker (Eds.),
  Self-Validating Numerics for Function Space Problems, Notes and Reports in
  Computer Science and Applied Mathematics, Academic Press, 1984, pp. 142--190.
\newblock \href
  {https://doi.org/https://doi.org/10.1016/B978-0-12-402020-7.50011-6}
  {\path{doi:https://doi.org/10.1016/B978-0-12-402020-7.50011-6}}.
\newline\urlprefix\url{https://www.sciencedirect.com/science/article/pii/B9780124020207500116}

\bibitem{1031925}
Y.~Cassivi, L.~Perregrini, P.~Arcioni, M.~Bressan, K.~Wu, G.~Conciauro,
  Dispersion characteristics of substrate integrated rectangular waveguide,
  IEEE Microwave and Wireless Components Letters 12~(9) (2002) 333--335.
\newblock \href {https://doi.org/10.1109/LMWC.2002.803188}
  {\path{doi:10.1109/LMWC.2002.803188}}.

\bibitem{1643580}
D.~Deslandes, K.~Wu, Accurate modeling, wave mechanisms, and design
  considerations of a substrate integrated waveguide, IEEE Transactions on
  Microwave Theory and Techniques 54~(6) (2006) 2516--2526.
\newblock \href {https://doi.org/10.1109/TMTT.2006.875807}
  {\path{doi:10.1109/TMTT.2006.875807}}.

\bibitem{article}
A.~Adabi, M.~Tayarani, Substrate integration of dual inductive post waveguide
  filter, Progress in Electromagnetics Research B 7 (2008) 321--329.
\newblock \href {https://doi.org/10.2528/PIERB08051002}
  {\path{doi:10.2528/PIERB08051002}}.

\bibitem{Salehian2023}
K.~Salehian, M.~Tayarani, \href{https://doi.org/10.1038/s41598-023-47490-1}{A
  novel siggw dual post band-pass filter for {5G} millimeter-wave band
  applications with a transmission zero}, Scientific Reports 13~(1) (2023)
  20743.
\newblock \href {https://doi.org/10.1038/s41598-023-47490-1}
  {\path{doi:10.1038/s41598-023-47490-1}}.
\newline\urlprefix\url{https://doi.org/10.1038/s41598-023-47490-1}

\bibitem{SEDIGHIMARAGHEH2019152885}
S.~{Sedighi Maragheh}, M.~Dousti, M.~Dolatshahi, B.~Ghalamkari,
  \href{https://www.sciencedirect.com/science/article/pii/S1434841119309501}{Tunable
  dual-band bandpass filter for multi-standard applications}, AEU -
  International Journal of Electronics and Communications 111 (2019) 152885.
\newblock \href {https://doi.org/https://doi.org/10.1016/j.aeue.2019.152885}
  {\path{doi:https://doi.org/10.1016/j.aeue.2019.152885}}.
\newline\urlprefix\url{https://www.sciencedirect.com/science/article/pii/S1434841119309501}

\bibitem{WU2024154975}
L.~Wu, Y.~Wu, X.~Cheng, R.~Huang, W.~Wang,
  \href{https://www.sciencedirect.com/science/article/pii/S1434841123004491}{Design
  of dual-mode dual-band filter based on multilayer groove gap waveguide}, AEU
  - International Journal of Electronics and Communications 173 (2024) 154975.
\newblock \href {https://doi.org/https://doi.org/10.1016/j.aeue.2023.154975}
  {\path{doi:https://doi.org/10.1016/j.aeue.2023.154975}}.
\newline\urlprefix\url{https://www.sciencedirect.com/science/article/pii/S1434841123004491}

\bibitem{10379032}
A.~Zhu, Z.~Li, L.~Cheng, C.~Hu, R.~N. Mahapatra, An amplitude tunable dual-band
  bandpass filter with perfect absorption and its sensing applications, IEEE
  Sensors Journal 24~(4) (2024) 4387--4399.
\newblock \href {https://doi.org/10.1109/JSEN.2023.3346441}
  {\path{doi:10.1109/JSEN.2023.3346441}}.

\bibitem{doi:10.1049/PBEW021E}
N.~Marcuvitz,
  \href{https://digital-library.theiet.org/doi/abs/10.1049/PBEW021E}{Waveguide
  Handbook}, The Institution of Engineering and Technology, 1986.
\newblock \href
  {http://arxiv.org/abs/https://digital-library.theiet.org/doi/pdf/10.1049/PBEW021E}
  {\path{arXiv:https://digital-library.theiet.org/doi/pdf/10.1049/PBEW021E}},
  \href {https://doi.org/10.1049/PBEW021E} {\path{doi:10.1049/PBEW021E}}.
\newline\urlprefix\url{https://digital-library.theiet.org/doi/abs/10.1049/PBEW021E}

\end{thebibliography}

\end{document}